\documentclass[10pt,twocolumn,letterpaper]{article}

\makeatletter
\@namedef{ver@everyshi.sty}{}
\makeatother
\usepackage{tikz}
\usepackage[pagenumbers]{cvpr} %

\setcitestyle{square, comma, numbers,sort&compress, super}
\usepackage{enumitem}
\usepackage{algpseudocode}
\usepackage[font=small,labelfont=bf]{caption}
\usepackage{array}
\usepackage{multirow}
\usepackage{booktabs}
\usepackage{algorithm}
\usepackage{subcaption}
\usepackage[normalem]{ulem}
\usepackage{xparse}
\usepackage{pifont}
\usepackage{bm}
\usepackage{threeparttable}
\usepackage{etoolbox}

\usepackage{listings}
\usepackage{mwe} %
\usepackage{makecell}
\usepackage{color, colortbl}
\usepackage{tabularx}
\usepackage{pifont}%
\usepackage[accsupp]{axessibility}

\usepackage[scaled=0.85]{DejaVuSansMono}

\definecolor{citecolor}{HTML}{0071bc}
\usepackage[breaklinks=true,bookmarks=false,colorlinks,bookmarks=false, citecolor=citecolor, pagebackref]{hyperref}

\usepackage{graphicx}
\usepackage{amsmath}
\usepackage{amssymb}
\usepackage{pgfplots}

\pgfplotsset{compat=1.16}
\usepgfplotslibrary{groupplots}
\usetikzlibrary{patterns}
\usetikzlibrary{plotmarks}

\newlength\savewidth\newcommand\shline{\noalign{\global\savewidth\arrayrulewidth
  \global\arrayrulewidth 1pt}\hline\noalign{\global\arrayrulewidth\savewidth}}

\newlength\thinwidth

\definecolor{Gray}{gray}{0.92}
\definecolor{DarkGray}{gray}{0.5}
\newcolumntype{x}{>{\columncolor{Gray}}c}
\newcolumntype{H}{>{\setbox0=\hbox\bgroup}c<{\egroup}@{}}
\definecolor{LightCyan}{rgb}{0.88,1,1}
\definecolor{altRowColor}{gray}{0.92}
\definecolor{highlightRowColor}{rgb}{0.9, 0.9, 1}

\newcommand{\colorcell}{\cellcolor{Gray}}

\definecolor{GrayNumber}{gray}{0.5}
\definecolor{GrayXMark}{gray}{0.7}

\newcommand{\OURS}{{\textsc{Aim}}\xspace}
\newcommand{\Ours}{\OURS}

\newcommand{\DFN}{DFN-2B+\xspace}

\newcommand{\tablestyle}[2]{\setlength{\tabcolsep}{#1}\renewcommand{\arraystretch}{#2}\centering\footnotesize}

\expandafter\def\expandafter\normalsize\expandafter{%
    \normalsize%
    \setlength\abovedisplayskip{2pt}%
    \setlength\belowdisplayskip{5pt}%
    \setlength\abovedisplayshortskip{-5pt}%
    \setlength\belowdisplayshortskip{2pt}%
}

\usepackage[capitalize]{cleveref}
\crefname{section}{\S}{\S\S}
\crefname{subsection}{\S}{\S\S}
\crefformat{table}{Table~#2#1#3}
\crefformat{figure}{Figure~#2#1#3}
\crefformat{equation}{Eq~#2#1#3}

\title{Scalable Pre-training of Large Autoregressive Image Models}

\author{
 \scalebox{0.85}{Alaaeldin El-Nouby \quad \quad Michal Klein \quad \quad Shuangfei Zhai \quad \quad  Miguel Angel Bautista} \\
  \scalebox{0.85}{Alexander Toshev \quad \quad Vaishaal Shankar \quad \quad Joshua M Susskind \quad \quad Armand Joulin${^{*}}$} \\ [0.2cm]
  Apple \\
  {\small \url{https://github.com/apple/ml-aim}}
}

\begin{document}

\twocolumn[{%
\renewcommand\twocolumn[1][]{#1}%
\maketitle
\begin{center}
    \centering
    \captionsetup{type=figure}
    \begin{subfigure}[t]{0.48\linewidth}
        \definecolor{CustomBlue}{rgb}{0.11764705882352941, 0.5647058823529412, 1.0}

\begin{tikzpicture}
    \begin{axis}[
        xtick={0.3169, 0.3119, 0.3084, 0.3039, 0.2989}, %
        legend pos=south east,
        xticklabels={0.3169, 0.3119, 0.3084, 0.3039, 0.2989}, %
        xmin=0.298,
        xmax=0.3155,
        ymin=75.1,
        x dir=reverse,
        grid=both,
        grid style={line width=.1pt, draw=gray!10},
        major grid style={line width=.2pt,draw=gray!50},
        minor tick num=2,
        axis x line*=bottom,
        axis y line*=left,
        height=1.8in,
        width=\linewidth,
        ylabel style= {align=center, font=\small},
        xlabel style = {font=\small},
        ylabel={Top-1 accuracy \\ \scalebox{0.7}{(15 benchmarks)}},
        xlabel={Pre-training validation loss \scriptsize{(IN-1k)}},
        yticklabel style = {font=\small},
        xticklabel style = {font=\small},
        legend style={cells={align=left}, font=\footnotesize},
        legend cell align={left},
        nodes near coords,
        nodes near coords style={
                anchor=south east,
                font=\footnotesize,
                color=black,
            },
        point meta=explicit symbolic,
    ]

    \addplot[mark=pentagon*, CustomBlue, mark options={solid}, line width=1.5pt, mark size=3.0pt] plot coordinates {
        (0.3119, 75.6) [\Ours-0.6B]
    };
    \addplot[mark=triangle*, CustomBlue, mark options={solid}, line width=1.5pt, mark size=3.0pt] plot coordinates {
        (0.3084,  76.1) [\Ours-1B]
        
    };
    \addplot[mark=oplus*, CustomBlue, mark options={solid}, line width=1.0pt, mark size=3.0pt] plot coordinates {
        (0.3039,  77.1) [\Ours-3B]
    };

    \addplot[mark=square*, CustomBlue, mark options={solid}, line width=1.0pt, mark size=3.0pt] plot coordinates {
        (0.2989,  78.6) [\Ours-7B]
    };
    
    \addplot[semithick, CustomBlue, line width=1.5pt] plot coordinates {
        (0.3119, 75.6)
        (0.3084, 76.1)
        (0.3039, 77.1)
        (0.2989, 78.6)
    };

    \end{axis}
\end{tikzpicture}
    \end{subfigure}
    \hfill
    \begin{subfigure}[t]{0.48\linewidth}
        \definecolor{CustomOrange}{rgb}{0.8823529411,0.63725490196,0.0156862745}

\begin{tikzpicture}
    \begin{axis}[
        xtick={1, 100, 2000}, %
        legend pos=south east,
        xticklabels={1M, 100M, 2B}, %
        xmode=log,
        grid=both,
        grid style={line width=.1pt, draw=gray!10},
        major grid style={line width=.2pt,draw=gray!50},
        minor tick num=2,
        axis x line*=bottom,
        axis y line*=left,
        height=1.8in,
        width=\linewidth,
        ylabel style= {align=center, font=\small},
        xlabel style = {font=\small},
        ylabel={Top-1 accuracy \\ \scalebox{0.7}{(15 benchmarks)}},
        xlabel={Number of unique images seen \scriptsize{(log scale)}},
        yticklabel style = {font=\small},
        xticklabel style = {font=\small},
        legend style={cells={align=left}, font=\footnotesize},
    ]

    \addplot[mark=pentagon*,  very thick, CustomOrange, mark options={solid}, line width=1.5pt, mark size=2.5pt] plot coordinates {
        (1,  72.8)
        (100,  73.8)
        (2000,  74.5)

    };
    \addlegendentry{\OURS-0.6B}

    \end{axis}
\end{tikzpicture}
    \end{subfigure}
    \caption{\textbf{\Ours scaling behavior} (Left) As we scale the capacity of \Ours, we observe improved
    performance for the pre-training objective which directly correlates with stronger downstream performance. (Right)
    \Ours exhibits stronger downstream performance when trained using larger sets of uncurated web
    data~\cite{gadre2023datacomp,fang2023data}. The downstream performance is the average attentive probe top-1 accuracy over a diverse
    set of \textbf{15 image recognition benchmarks}. All models are trained for the same number of updates.}
    \label{fig:teaser_figure}
\end{center}
}]
\makeatletter{\renewcommand*{\@makefnmark}{}
\let\thefootnote\relax\footnote{$^{*}$\scriptsize{Work done while at Apple. Now at Google
DeepMind.}}

\vspace{-3mm}
\begin{abstract}
This paper introduces \Ours, a collection of vision models pre-trained with an
autoregressive objective. 
These models are inspired by their textual counterparts, i.e., Large Language Models (LLMs), and exhibit similar scaling properties.
Specifically, we highlight two key findings: (1) the performance of the visual features scale with both the model capacity and the quantity of data,
(2) the value of the objective function correlates with the performance of the model on downstream tasks. 
We illustrate the practical implication of these findings by pre-training a 7 billion parameter \Ours on 2 billion images that achieves 84.0\% on ImageNet-1k with a frozen trunk. 
Interestingly, even at this scale, we observe no sign of saturation in performance, suggesting that \Ours potentially represents a new frontier for
training large-scale vision models.
The pre-training of \Ours is similar to the pre-training of LLMs, and does not require any image-specific strategy to stabilize the training at scale.
\end{abstract}
 \vspace{-7mm}
\section{Introduction}
\label{sec:intro}

Pre-training task agnostic models has become the standard in Natural Language
Processing with the recent revolution of large language models
(LLMs)~\citep{radford2019language, brown2020language, touvron2023llama}. These
models can solve complex reasoning tasks from a few
examples~\cite{brown2020language}, follow
instructions~\cite{ouyang2022training}, and now serve as the engine of widely
used AI assistants such as ChatGPT. A key factor contributing to their success
is the ability to consistently improve as the capacity (\ie, number of
parameters) or the amount of pre-training data~\cite{radford2019language}
increases. 

The scaling behavior of these models is remarkable for two key reasons. First,
even though these models are trained with a simple objective -- predicting the
next word in a sentence given its past -- they are able to learn  intricate
patterns over long contexts. Second, the scalability of this autoregressive
objective is mostly observed when used in conjunction with certain
architectures, and in particular Transformers~\cite{vaswani2017attention},
highlighting the potential synergy between the autoregressive pre-training and
this architecture. %

These observations naturally raise the follow-up question of whether the success
of scaling Transformers with an autoregressive objective is exclusive to text.
This is particularly significant considering that none of the aforementioned
elements are inherently specific to language modeling. Autoregressive objectives
take their roots in the data compression
literature~\cite{shannon1951prediction}, and similar approaches have been
investigated in audio~\cite{oord2018representation} and
images~\citep{van2016conditional, chen2020generative}. The Transformer
architecture has also been successfully used in other domains, in particular,
computer vision with the success of the Vision Transformers~(ViT)
\citep{dosovitskiy2020image}. Therefore, as a first step towards generalizing
the findings of LLMs, we explore if training ViT models with an autoregressive
objective leads to competitive performance, in terms of learning
representations, with the same scaling ability as LLMs.

In this paper, we introduce Autoregressive Image Models (\textbf{\Ours}), an
autoregressive approach for large-scale pre-training for visual features. We
revisit prior work in autoregressive representation learning such as
iGPT~\cite{chen2020generative} using a modern toolset that includes vision
transformers, collections of large-scale web
data~\cite{gadre2023datacomp,fang2023data} and recent advances in LLM
pre-training~\cite{touvron2023llama, hoffmann2022training}. Additionally, we
introduce two architectural modifications to adapt autoregressive pre-training
to visual features. First, instead of restricting the self-attention to be fully
causal as is typically the case for LLMs, we adopt a prefix attention, as in
T5~\cite{roberts2019exploring}. This choice enables moving to a fully
bidirectional attention during downstream tasks. Second, we use a heavily
parameterized token-level prediction head, inspired by the heads used in
contrastive learning~\cite{chen2020simple}. We observe that this modification
significantly improves the quality of the subsequent features with little
overhead during training. Overall, the training of \Ours is similar to the
training of recent LLMs and does not rely on any stability-inducing
techniques~\cite{touvron2021going, huang2016deep, dehghani2023scaling} that
supervised~\cite{touvron2021going, dehghani2023scaling} or
self-supervised~\cite{bao2021beit,oquab2023dinov2} methods need. 

We provide a study of a series of models, ranging from 600M to 7B parameters
pre-trained using 2B uncurated images with permissive licenses. Our \Ours models
exhibit strong scaling behavior w.r.t. the model size as shown in
Figure~\ref{fig:teaser_figure} where higher capacity models achieve better
downstream performance, measured as the average accuracy over 15 image
recognition benchmarks. More importantly, there is a correlation between the
value of our objective function on a validation set and the quality of the
subsequent frozen features. This observation confirms that the autoregressive
objective is adequate for the training of visual features. Furthermore, we
observe consistent improvement in downstream performance as we train on more
images, with no sign of saturation. Overall, these observations
are aligned with the previous studies on scaling large language models.

\section{Related Work}

\par \noindent \textbf{Autoregressive models.} While most of the literature on
autoregressive models come from language modeling~\cite{mikolovrecurrent,
bengio2000neural,radford2019language} or speech~\cite{oord2018representation,
oord2016wavenet}, few works have explored the potential of this approach for
images~\cite{larochelle2011neural, parmar2018image, van2016conditional,
salimans2017pixelcnn, chen2020generative, parmar2018image}. Of particular
interest,~\citet{van2016conditional} show that using an architecture adapted to
images, \eg, a convolution network, significantly improved over autoregressive
models built with more generic architecture~\cite{van2016pixel}, \eg, a
recurrent network~\cite{elman1990finding}.~\citet{parmar2018image} further
improve the quality of these autoregressive models by adopting the transformer
architecture~\cite{vaswani2017attention}. More
recently,~\citet{chen2020generative} have shown that scaling with more compute
leads to continuous improvements. Our work follows this line of research, and we
benefit from training on significantly more  data, and further improvement in
architecture design~\cite{dosovitskiy2020image},
training~\cite{touvron2021training,touvron2023llama} and understanding of the
scaling law~\cite{hoffmann2022training}. Concurrent to our
work,~\citet{bai2023sequential} demonstrate the effectiveness of large-scale
autoregressive vision models for in-context pixel prediction tasks (\eg,
semantic segmentation, depth estimation).

\par \noindent \textbf{Self-supervised pre-training.} Pre-training vision models
on datasets of images without supervision has been a fruitful area of research
in recent years~\cite{doersch2015unsupervised,misra2020self,zhang2016colorful,
bojanowski2017unsupervised,gidaris2018unsupervised,zhou2021ibot,
dosovitskiy2014discriminative}. Different approaches have been employed,
focusing on various proxy tasks for feature learning. For
example,~\citet{noroozi2016unsupervised} learn to re-arrange the order of
shuffled image patches. Some other works have relied on
clustering~\citep{bautista2016cliquecnn, caron2018deep, yan2020cluster,
caron2021emerging}. Another popular approach involves the use of a contrastive
objective, that resembles predictive coding, where the objective is to identify each
image~\cite{chen2020simple, he2020momentum}. Most recent contrastive approaches
include DINO~\citep{oquab2023dinov2}, BYOL \cite{grill2020bootstrap} or
iBot~\cite{zhou2021ibot}. In a similar vein, some works have proposed predictive
approaches~\cite{assran2023self,bardes2022vicreg} or a form of feature
whitening~\cite{zbontar2021barlow}. Closer to our approach are works inspired by
BERT~\cite{devlin2018bert} where patches are masked and predicted with an
autoencoder in either their discrete~\citep{bao2021beit} or
pixel~\citep{he2021masked} form.

\par \noindent \textbf{Other generative pre-training.} Autoregressive modeling
is a form of generative modeling, and few other generative approaches have been
considered to learn visual features. The first category leverages some form of
autoencoding where the pretext task corresponds to some denoising task. For
instance, the noise can be salt-and-pepper~\cite{vincent2010stacked} or
masking~\cite{pathak2016context,bao2021beit}. Another line of work leverages
Generative Adversarial Networks~(GANs)~\cite{goodfellow2014generative}. Most
notably, BigGAN~\cite{brock2018large} trains a large GAN and re-uses the image
discriminator to produce image features. More recently, DiffMAE
\citep{wei2023diffusion} used diffusion models to learn image features.

\par \noindent \textbf{Pre-training at scale.} There are numerous works on
scaling the pre-training of visual features with no
supervision~\cite{oquab2023dinov2,singh2023effectiveness,tian2021divide,
goyal2019scaling, goyal2022vision, caron2019unsupervised}. The most salient work
in this area is DINOv2 where they produce the best self-supervised features by
scaling the iBot method~\cite{zhou2021ibot} on a private dataset of 142M images
and a 460M parameter model. The conclusion from this work is that a carefully
tuned contrastive method scales reasonably well, but they do not exhibit the
scaling law that we observe with language modeling. They also rely on an
intricate implementation of contrastive learning to avoid the pitfalls described
by~\citet{chen2021empirical}. In parallel,~\citet{singh2023effectiveness} study
the scaling of Masked Autoencoders~(MAE)~\cite{he2017mask}. While the study
focuses on a weakly-supervised setup, it does not showcase strong improvements
to the self-supervised pre-training as the data is scaled to billions of images.
In contrast, we observe a clear benefit of scale on the quality of
our features, even at a scale of a few billions of parameters and billions of
images.

\section{Pre-training Dataset}
\label{sec:data}
We pre-train our models on the DFN dataset introduced by~\citet{fang2023data}.
This dataset is composed of a larger collection of 12.8B image-text
pairs~\cite{gadre2023datacomp} filtered from Common Crawl. The data has been
pre-processed to remove NSFW content, blur faces, and reduce contamination by
deduplicating against the evaluation sets. A data filtering
network~\cite{fang2023data} ranks the samples in the 12.8B collection according
to the alignment score between images and their corresponding caption. A subset
of 2B images, called DFN-2B, has been extracted from the DataComp 12.8B dataset
~\cite{gadre2023datacomp} by keeping the top 15\% samples. Note that other than
the privacy and safety filters, this process does not include any additional
curation based on the image content. Since our pre-training does not require
text, our method could be pre-trained using larger image collections that are
not paired with captions or have low image-text alignment such as the rest of
DataComp 12.8B.

Motivated by the common practice in LLM pre-training~\cite{touvron2023llama} of
oversampling high-quality data sources such as Wikipedia and Books, during
pre-training, we sample images from DFN-2B with a probability of $p=0.8$ and
sample images from ImageNet-1k with a probability of $p=0.2$. We refer to such
dataset as \DFN.

\section{Approach}
\label{sec:approach}

\subsection{Training Objective}

Our training objective follows that of a standard autoregressive model applied
on a sequence of image patches. More precisely, an image $x$ is split into a
grid of $K$ non-overlapping patches $x_k$, $k\in[1, K]$, which collectively form
a sequence of tokens. We assume that the sequence order is fixed across all
images, and we use a raster (row-major) ordering by default unless otherwise
specified. Given the above order, the probability of an image can be factorized
as a product of patch conditional probabilities:
\begin{equation}
P(x) =  \prod_{k=1}^K P(x_k~|~x_{<k}),
\end{equation}
where $x_{<k}$ denotes the set of the first $k-1$ patches, and is the context
used to predict the $k^{\text{th}}$ patch. As opposed to language modeling, our sequences
have a fixed length of $K$ that fits in memory and hence we do not need to
truncate the context length. The training loss over a set $\mathcal{X}$ of
images is then defined as the negative log-likelihood (NLL):
\begin{equation*}
    \sum_{x\in\mathcal{X}} \sum_{k=1}^K -\log P(x_k~|~x_{<k}).
\end{equation*}

\begin{figure}[t!]
    \centering
    \includegraphics[width=0.9\linewidth]{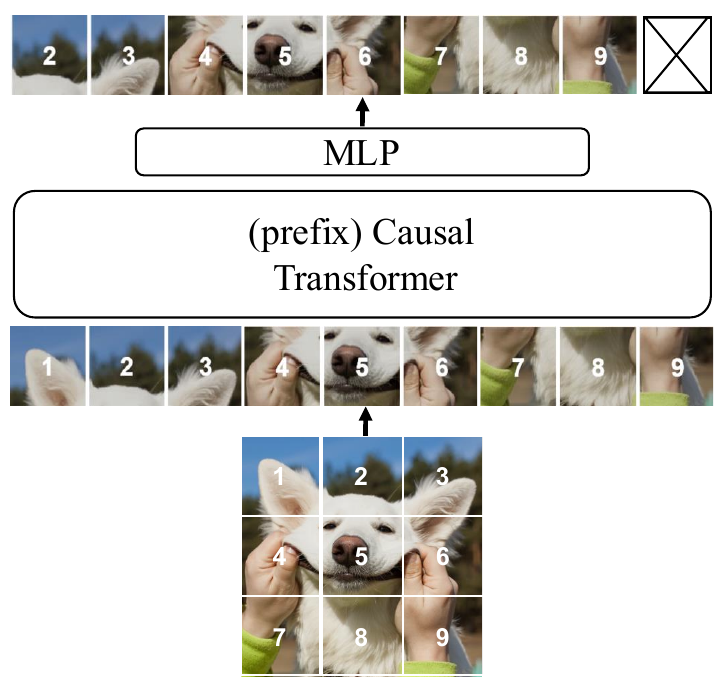}
    \caption{\textbf{\Ours pre-training overview.}. Input images are split into
    non-overlapping patches and embedded linearly
    following~\citet{dosovitskiy2020image}. The patch features are fed to a
    transformer in which the self-attention operation is causally masked to
    prevent attending to preceding positions. Afterward, a heavily parameterized
    MLP processes each of the patch features independently and finally projects
    it to pixel space. The targets correspond to the input sequence shifted one
    position to the left, requiring the model to predict the next patch in
    raster order.}
    \label{fig:aim_overview}
\end{figure}

Minimizing this objective over an infinite amount of images, with no further
assumptions, is theoretically equivalent to learning the true underlying image
distribution.

\par \noindent \textbf{Prediction loss}
Our training objective naturally gives rise to certain variants of losses, each
corresponding to a choice of the distribution $P(x_k~|~x_{<k})$. By default, we
adopt a normalized pixel-level regression loss similar to~\citet{he2021masked}. This loss corresponds to setting $P(x_k~|~x_{<k})$ as
Gaussian distributions with a constant variance. Namely, given
$\hat{x}_k(\theta)$ as the prediction of the $k^{\text{th}}$ patch from a
network parameterized with $\theta$, and $x_k$ as its corresponding ground-truth
value, our objective is to minimize the sum $\ell_2$ squared distance between
the prediction and the ground-truth:
\begin{equation}
\min_{\theta} \frac{1}{K}\sum_{k=1}^K \|\hat{x}_k(\theta)-x_k\|_2^2.
\end{equation}
We also consider a cross-entropy loss with patches converted to discrete tokens
using an offline tokenizer. Our ablation studies show that these designs work,
although they do not produce as strong features as the pixel-wise loss.

\begin{figure}[t!]
    \centering
    \includegraphics[width=.85\linewidth]{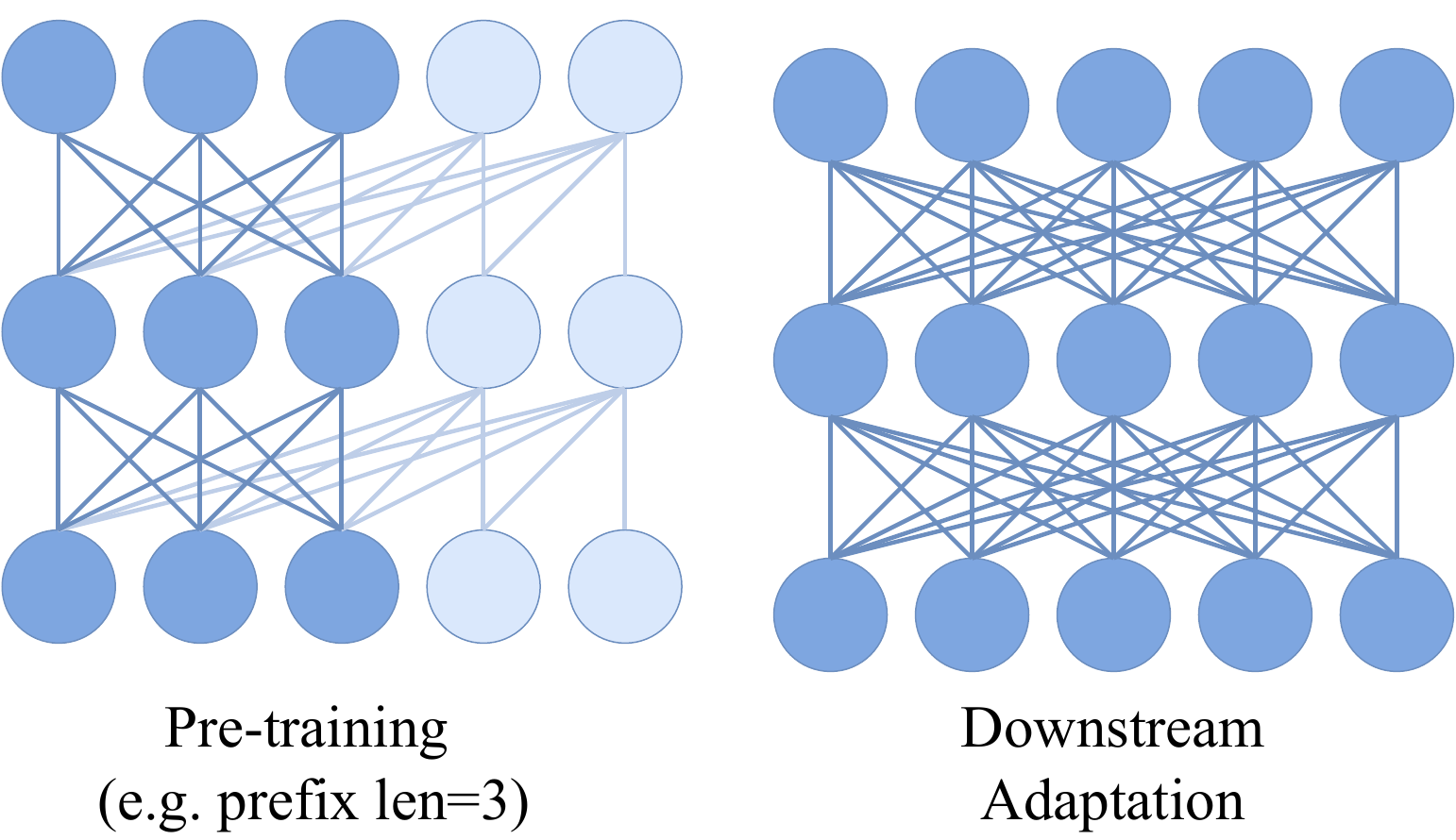}
    \caption{\textbf{Prefix causal attention.} During pre-training we uniformly
    sample a prefix length $S$. The attention for the first
    $S$ patches are set to be bidirectional and loss is only
    computed for the remaining patches in the image. During adaptation to
    downstream tasks, this allows us to drop the attention causal mask,
    improving the downstream performance.}
    \label{fig:prefix_attn}
\end{figure}

\begin{table}[t!]
    \centering
    \setlength{\tabcolsep}{3pt}
    \renewcommand{\arraystretch}{1.2}
    \resizebox{1.0\linewidth}{!}{
    \begin{tabular}{lccccccc}
        \textbf{Model} & \textbf{\#Params}  & \textbf{Hidden size} & \textbf{Layers} & \textbf{LR} &\textbf{ \#Patches} & \textbf{Batch size}\\
         \shline
         \Ours-0.6B  & 0.6B & 1536  & 24 & $1e^{-3}$ & 0.5T & 4096 \\
         \Ours-1B & 1.2B & 2048  & 24 & $1e^{-3}$ & 1.2T & 4096 \\
         \Ours-3B & 2.7B & 3072  & 24 & $1e^{-3}$ & 1.2T & 4096 \\
         \Ours-7B & 6.5B & 4096  & 32 & $1e^{-3}$ & 1.2T & 4096 \\
    \end{tabular}}
    \caption{\textbf{Model specifications.} We provide the embedding dimension, number of layers, and parameter count for all \Ours variants. We also provide the learning rate and batch size during pre-training. For \Ours with 1B parameters and higher, the pre-training process involves 1.2M iterations, which corresponds to 1.2 trillion patches, or 5B images, seen during pre-training.}
    \label{tab:model_specs}
\end{table}

\subsection{Architecture}

As the backbone, we adopt the Vision Transformer
architecture~(ViT)~\cite{dosovitskiy2014discriminative}. For scaling in the
model capacity, we follow the common practice in language modeling and we
prioritize expanding width rather than depth~\cite{radford2019language,
touvron2023llama}. In Table~\ref{tab:model_specs}, we provide an overview of the
design parameters of \Ours, including its depth and width, as well as the amount
of data and optimization scheme for each model capacity. The overall model is
illustrated in Figure \ref{fig:aim_overview}.

During pre-training, we apply causal masks to the self-attention layers to model
the probability of a patch given the preceding patches. More precisely, given a
self-attention layer, the embedding for the patch $i$ is computed by:
\begin{equation}
    y_i = \sum_{k=1}^K a_{ik} v_i,
\end{equation}
where $a_{ik}$ is the attention weight and $v_k$ the value embedding.  To
enforce the desired constraints, we utilize a causal mask for the attention
weights, where $a_{ik} = 0$ for $k > i$, and $\sum_{k=1}^K a_{ik} = 1$. This
approach enables us to process the image with a single forward pass during
training, without incurring additional computational overhead.

\par \noindent \textbf{Prefix Transformer.} The autoregressive objective in
pre-training requires a causal mask in the self-attention operation. However,
this differs from the standard usage of ViT models in downstream tasks, where
bidirectional self-attention is employed. This discrepancy leads to a decrease
in performance, irrespective of whether the causal mask is retained during
downstream adaptation or not (as shown in the ablations presented in
Table~\ref{tab:ablations}). To address this issue, we propose to consider the
initial patches of the sequence, referred to as the prefix, as a context for
predicting the remaining patches following the PrefixLM formulation of
\citet{raffel2020exploring}. The prefix patches are excluded from the
autoregressive prediction and therefore are not constrained to be causal. More
precisely, we select a prefix length of size $S\in [1, K-1]$, and remove the
causal mask, \ie, $a_{i, k} > 0$ for $k < S$. This modification helps the model
to work in the absence of causal masking, allowing it to be removed during
downstream adaptation. This approach improves the performance of the model in
downstream tasks and eliminates the need for architectural changes to ViT.
Figure~\ref{fig:prefix_attn} illustrates the difference between causal and
prefix attention.

\par \noindent \textbf{MLP prediction heads.} It is a common practice to adopt
certain prediction heads during pre-training, which are discarded when
transferring to downstream tasks~\cite{chen2020simple, chen2021empirical,
caron2020unsupervised, caron2021emerging, grill2020bootstrap}.  The purpose of
these heads is to prevent the trunk features from becoming too specialized in
the pre-training objective, thus enhancing their suitability for downstream
transfer. We opt for a simple design where we use $N$ blocks of MLP on top of
the final transformer layer, processing each patch independently. We observed
that this design strikes a good balance between performance and the additional
costs incurred during pre-training.

\begin{figure*}[t!]
    \centering
    \captionsetup{type=figure}
    \begin{subfigure}[b]{0.49\linewidth}
        \input{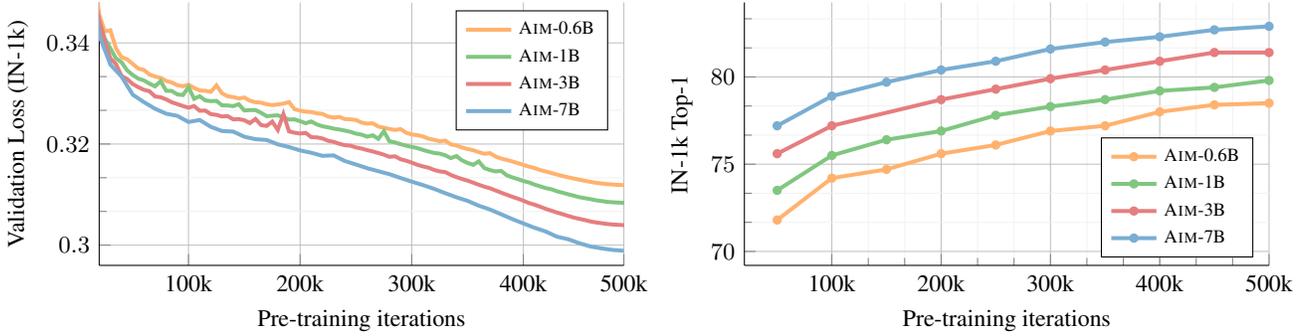}
    \end{subfigure}
    \hfill
    \begin{subfigure}[b]{0.49\linewidth}
        \definecolor{CustomOrange}{rgb}{0.8823529411,0.63725490196,0.0156862745}
\definecolor{darkspringgreen}{rgb}{0.09, 0.45, 0.27}
\definecolor{colorA}{HTML}{1F77B4} %
\definecolor{colorB}{HTML}{FF7F0E} %
\definecolor{colorC}{HTML}{2CA02C} %
\definecolor{colorD}{HTML}{D62728} %

\begin{tikzpicture}
    \begin{axis}[
        xtick={100000, 200000, 300000, 400000, 500000}, %
        legend pos=south east,
        xticklabels={100k, 200k, 300k, 400k, 500k}, %
        xmin=20000,
        xmax=500000,
        ymin=69.2,
        grid=both,
        grid style={line width=.1pt, draw=gray!10},
        major grid style={line width=.2pt,draw=gray!50},
        minor tick num=2,
        axis x line*=bottom,
        axis y line*=left,
        height=2in,
        width=\linewidth,
        ylabel style= {align=center, font=\small},
        xlabel style = {font=\small},
        ylabel={IN-1k Top-1},
        xlabel={Pre-training iterations},
        yticklabel style = {font=\small},
        xticklabel style = {font=\small},
        legend style={cells={align=left}, font=\scriptsize},
        legend cell align={left},
        scaled ticks=false,
    ]
    \addplot[colorB!60, mark=*, mark options={solid}, line width=1.5pt, mark size=1.0pt] plot coordinates {
        (50000, 71.8)    
        (100000, 74.2)    
        (150000, 74.7)    
        (200000, 75.6)    
        (250000, 76.1)    
        (300000, 76.9)    
        (350000, 77.2)    
        (400000, 78.0)    
        (450000, 78.4)    
        (500000, 78.5)    
    };
    \addlegendentry{\OURS-0.6B}
    \addplot[colorC!60, mark=*, mark options={solid}, line width=1.5pt, mark size=1.0pt] plot coordinates {
        (50000, 73.5)    
        (100000, 75.5)    
        (150000, 76.4)    
        (200000, 76.9)    
        (250000, 77.8)    
        (300000, 78.3)    
        (350000, 78.7)    
        (400000, 79.2)    
        (450000, 79.4)    
        (500000, 79.8)    
    };
    \addlegendentry{\OURS-1B}
    
    \addplot[colorD!60, mark=*, mark options={solid}, line width=1.5pt, mark size=1.0pt] plot coordinates {
        (50000, 75.6)    
        (100000, 77.2)    
        (200000, 78.7)    
        (250000, 79.3)    
        (300000, 79.9)    
        (350000, 80.4)    
        (400000, 80.9)    
        (450000, 81.4)    
        (500000, 81.4)    
    };
    \addlegendentry{\OURS-3B}

    \addplot[colorA!60, mark=*, mark options={solid}, line width=1.5pt, mark size=1.0pt] plot coordinates {
        (50000, 77.2)    
        (100000, 78.9)    
        (150000, 79.7)    
        (200000, 80.4)    
        (250000, 80.9)    
        (300000, 81.6)    
        (350000, 82.0)    
        (400000, 82.3)    
        (450000, 82.7)    
        (500000, 82.9)    
    };
    \addlegendentry{\OURS-7B}
    
    \end{axis}
\end{tikzpicture}
    \end{subfigure}
    \vspace{-2mm}
    \caption{\textbf{\Ours pre-training across model sizes.} We observe a clear
    improvement in the performance of the pre-training objective with increasing
    the capacity of \Ours. Moreover, the downstream performance
    (IN-1k top-1) is monotonically improving for higher capacity models as well
    as with longer pre-training. We do not observe clear signs of plateauing
    during pre-training even after training for 500k iterations, indicating that
    \Ours can benefit from even longer pre-training schedules. Note
    that the loss saturation at the very end of training is caused by the
    cosine decay schedule where the learning rate is effectively zero.}
    \label{fig:pretrain_loss_acc}
\end{figure*}

\par \noindent \textbf{Straightforward implementation.} It is worth noting that
\Ours does not require particular optimization stability-inducing mechanisms
such as LayerScale~\cite{touvron2021going}, stochastic
depth~\cite{huang2016deep}, QK-Norm~\cite{dehghani2023scaling}, or freezing the
patch projector~\cite{chen2021empirical}. These mechanisms have been crucial for
the success of other methods, either supervised or self-supervised. On the
contrary, we observe that \Ours scales using the same set of optimization
hyperparameters across model sizes with no further tuning (see
Table~\ref{tab:model_specs}).

We add sinusoidal positional embeddings~\cite{vaswani2017attention} to the input
patches before the transformer and before the MLP head.  We use a standard
expansion ratio of 4 for all the MLP blocks in the trunk and the head. We drop
the bias term for simplicity, and unlike the original ViT, we do not append a
classification token to the input. By default, we use 12 blocks for the MLP head
for all model capacities. The pixel targets are normalized per patch before the
loss computation following~\citet{he2021masked}.  We train our model using
\texttt{bfloat16} precision. We use the AdamW~\cite{loshchilov2017decoupled}
optimizer with linear warmup and a cosine decay schedule.  We detail the
hyperparameters used for pre-training and downstream adaptation in
Appendix~\ref{app:hparams}.

\par \noindent \textbf{Downstream adaptation.} Pre-training large-scale models
is a resource-intensive process, and even fine-tuning them is demanding.
Consequently, we focus on scenarios where all model weights are fixed for
downstream tasks. In this context, we only train a classification head,
which mitigates the risk of overfitting on small downstream datasets and
significantly reduces the adaptation cost.

Unlike contrastive learning, our loss is computed independently for each patch.
This means that our pre-training does not incorporate any notion of global image
descriptors, and hence, we do not have any image level token. While some methods
rely on global average pooling to build a global feature from the patch
features, we find that our approach, along with other generative approaches like
MAE, benefit more from an attention pooling operation~\cite{lee2019set} placed
before the linear classifier. Other works \cite{yu2022coca, touvron2021going,
anonymous2023vjepa} have adopted this attention pooling to improve performance
with minimal increase in parameters or FLOPs.

Specifically, given a set of patch features~$P = \{p_{i} \mid 1 \leq i \leq
K\}$, we compute a global descriptor $\hat{p}$ through multi-head attention
pooling over the patch features as:
\begin{equation}    
    \hat{p_h} = \sum^{K}_{i=1} \frac{\text{exp}(q_h^{T} W^{k}_h p_{i})}{\sum_{j=1}^K\text{exp}(q_h^{T} W^{k}_h p_{j})} W^{v}_h p_{i},
\end{equation}
\noindent where for each attention head $h = \{1,...,H\}$, $W^{k}_h, W^{v}_h \in
R^{d_h \times d}$ correspond to the key and value weight matrices, respectively;
$q_h$ is a learnable query vector. And we obtain the pooled feature as $\hat{p}
= [p_1,..., p_H], \hat{p} \in R^d$, which serves as the input to the linear
classifier. By default, we set the number of heads $H= \frac{d}{d_h}$, which
makes the total number of learnable parameters $2d^2 + d$, a negligible cost
compared to the main model size. Including this attention pooling makes the
entire operation not strictly linear, and, therefore we refer to it as
``Attentive Probe''. Nevertheless, the advantages of linear probing, e.g., low
additional parameter count and a reduced risk of overfitting, are preserved with
this probe.

\section{Results}

\subsection{Impact of scaling}

We measure the impact when scaling our approach in terms of parameters and
training data. In particular, we investigate whether there is a correlation
between the pre-training objective and the downstream performance across
benchmarks. We also look at the effect of scaling on the value of the loss
function. For all of these experiments, we report the value of our loss function
on the validation set of IN-1k. 

\par \noindent \textbf{Loss and performance during training.} In
Figure~\ref{fig:pretrain_loss_acc}, we measure for each model the value of the
pre-training loss and the classification accuracy on the validations set, as a
function of the number of training iterations. We observe that both probes
improve accordingly during the entire training, showing that optimizing our
objective directly results in better downstream performance.

\par \noindent \textbf{Number of parameters.} We observe that the loss value and
the accuracy of the downstream task improve as we scale the capacity of our
models. This observation is consistent with the trend observed in LLMs and can
be directly attributed to the optimization of our objective function, which in
turn leads to the learning of stronger representations.

\par \noindent \textbf{Number of images.} In Figure~\ref{fig:data_mix_loss}, we
show the progression of the validation loss as we pre-train on either a small
curated dataset of 1M images, \ie, IN-1k, or a larger set of 2B images, \ie
\DFN. It is not surprising that training on IN-1k leads rapidly to a low
validation loss as measured on the same distribution. However, this loss
deteriorates at the end of the training, indicating an overfitting to the
training data. When training on the uncurated DFN-2B dataset, the model starts
from a higher validation loss but the loss continues to decrease with no sign of
overfitting. When the same dataset is augmented with a small amount of IN-1k
data, as detailed in~\cref{sec:data}, we observe further improvement in the
performance that eventually surpasses pre-training on IN-1k. We confirm that the
resulting model also leads to a better downstream performance in
~\cref{tab:datset_mix_downstream}. 

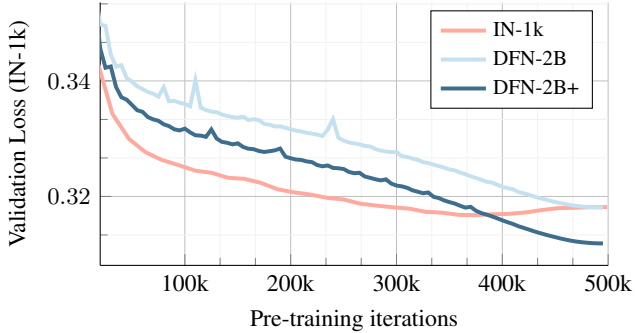
\begin{figure}[t!]
    \centering
    \definecolor{CustomOrange}{rgb}{0.8823529411,0.63725490196,0.0156862745}
\definecolor{darkspringgreen}{rgb}{0.09, 0.45, 0.27}
\definecolor{bittersweet}{rgb}{1.0, 0.44, 0.37}
\definecolor{darkpurple}{rgb}{.5,.0,.5}

\definecolor{colorA}{HTML}{C5E1EF}
\definecolor{colorB}{HTML}{6CB0D6}
\definecolor{colorC}{HTML}{0D4A70}

\begin{tikzpicture}
    \begin{axis}[
        xtick={100000, 200000, 300000, 400000, 500000}, %
        legend pos=north east,
        xticklabels={100k, 200k, 300k, 400k, 500k}, %
        xmin=20000,
        xmax=500000,
        grid=both,
        grid style={line width=.1pt, draw=gray!10},
        major grid style={line width=.2pt,draw=gray!50},
        minor tick num=2,
        axis x line*=bottom,
        axis y line*=left,
        height=2in,
        width=\linewidth,
        ylabel style= {align=center, font=\small},
        xlabel style = {font=\small},
        ylabel={Validation Loss (IN-1k)},
        xlabel={Pre-training iterations},
        yticklabel style = {font=\small},
        xticklabel style = {font=\small},
        legend style={cells={align=left}, font=\footnotesize},
        scaled ticks=false,
        legend cell align={left},
    ]

    \addplot[bittersweet!60, mark options={solid}, line width=1.5pt, mark size=0.2pt] plot coordinates {
        (312,0.601072630727291)
        (15913,0.344284168322086)
        (31514,0.334325209103822)
        (47115,0.329975155172348)
        (62716,0.327667201050519)
        (78317,0.326310546893477)
        (93918,0.325434265316724)
        (109519,0.324479437807798)
        (125120,0.324098078321218)
        (140721,0.323235522900819)
        (156322,0.323043500753045)
        (171923,0.322343730385303)
        (187524,0.321247319724559)
        (203125,0.320711087941527)
        (218726,0.320326918793916)
        (234327,0.319764519311189)
        (249928,0.319476262823939)
        (265529,0.31875854306519)
        (281130,0.318409268223643)
        (296731,0.31812781437397)
        (312332,0.317879683211445)
        (327933,0.317334144406914)
        (343534,0.317183809127807)
        (359135,0.31680183032453)
        (374736,0.316762300645709)
        (390337,0.316913466354608)
        (405938,0.316980764741301)
        (421539,0.317218131087422)
        (437140,0.317641621708273)
        (452741,0.317978748685121)
        (468342,0.318035283569097)
        (483943,0.318151690143346)
        (499232,0.318158558310866)
    };
    \addlegendentry{IN-1k}
    \addplot[colorA, mark options={solid}, line width=1.5pt, mark size=0.2pt] plot coordinates {
        (5001,0.385586066007614)
        (10002,0.364854836492538)
        (15003,0.355747437410354)
        (20004,0.349791505393981)
        (25005,0.349509607954025)
        (30006,0.344521473464965)
        (35007,0.342451424808502)
        (40008,0.342718098754882)
        (45009,0.340493864231109)
        (50010,0.339872300910949)
        (55011,0.339054943647384)
        (60012,0.338591733627319)
        (65013,0.338088690986633)
        (70014,0.33778790230751)
        (75015,0.337273787145614)
        (80016,0.338897430677413)
        (85017,0.336475707578659)
        (90018,0.336548719825744)
        (95019,0.336223309164047)
        (100020,0.336007915773391)
        (105021,0.335616840391159)
        (110022,0.340076233501434)
        (115023,0.335361175765991)
        (120024,0.334946382703781)
        (125025,0.334673883962631)
        (130026,0.334598901615142)
        (135027,0.334175494871139)
        (140028,0.333932119913101)
        (145029,0.33373977923423)
        (150030,0.333624319448471)
        (155031,0.333812945213317)
        (160032,0.333209450817108)
        (165033,0.333367110671997)
        (170034,0.333208644199371)
        (175035,0.33255685090065)
        (180036,0.332432412176132)
        (185037,0.332111950540542)
        (190038,0.332134452695846)
        (195039,0.331884400539398)
        (200040,0.331645318174362)
        (205041,0.331389213724136)
        (210042,0.331245918111801)
        (215043,0.331017067651748)
        (220044,0.33087904015541)
        (225045,0.330700648784637)
        (230046,0.330459929647445)
        (235047,0.331584476785659)
        (240048,0.333386854095459)
        (245049,0.33016597743988)
        (250050,0.329659350481033)
        (255051,0.329350914001464)
        (260052,0.32917911028862)
        (265053,0.329025646057128)
        (270054,0.328755994291305)
        (275055,0.328656963090896)
        (280056,0.328195681362152)
        (285057,0.327940779428482)
        (290058,0.327768778505325)
        (295059,0.327685610542297)
        (300060,0.32766433057785)
        (305061,0.326983192014694)
        (310062,0.326819861040115)
        (315063,0.326558480281829)
        (320064,0.326338492727279)
        (325065,0.325934539661407)
        (330066,0.325775241556167)
        (335067,0.325506675596237)
        (340068,0.325246111402511)
        (345069,0.324987473902702)
        (350070,0.324668722324371)
        (355071,0.324507764487266)
        (360072,0.324141015453338)
        (365073,0.32384560183525)
        (370074,0.32348897149086)
        (375075,0.323199627804756)
        (380076,0.322882405848503)
        (385077,0.322541429166793)
        (390078,0.322412911539077)
        (395079,0.322033242101669)
        (400080,0.321675621318817)
        (405081,0.32142385892868)
        (410082,0.321255428056716)
        (415083,0.320778018283844)
        (420084,0.320544202098846)
        (425085,0.320294805288314)
        (430086,0.319991654448509)
        (435087,0.319718899655342)
        (440088,0.319492250723838)
        (445089,0.319254287157058)
        (450090,0.31902173669815)
        (455091,0.318836885170936)
        (460092,0.318666192021369)
        (465093,0.31847513226509)
        (470094,0.318367791290283)
        (475095,0.318256612548828)
        (480096,0.318177197957038)
        (485097,0.318104471020698)
        (490098,0.318078814072608)
        (495099,0.318058552169799)
    };
    \addlegendentry{DFN-2B}
    \addplot[colorC!80, mark options={solid}, line width=1.5pt, mark size=0.2pt] plot coordinates {
        (5001,0.379782930011749)
        (10002,0.368867099666595)
        (15003,0.354315392026901)
        (20004,0.345615336503982)
        (25005,0.342272347469329)
        (30006,0.342492079601287)
        (35007,0.338935468072891)
        (40008,0.33717096253395)
        (45009,0.336753661842346)
        (50010,0.335842792491912)
        (55011,0.334914555578231)
        (60012,0.33464796136856)
        (65013,0.333747369556427)
        (70014,0.333341662034988)
        (75015,0.333099012975692)
        (80016,0.332485592470169)
        (85017,0.332173402013778)
        (90018,0.331623362188339)
        (95019,0.331425791206359)
        (100020,0.331738626270294)
        (105021,0.331090031166076)
        (110022,0.330586046371459)
        (115023,0.330528353109359)
        (120024,0.330324150791168)
        (125025,0.331594567461013)
        (130026,0.330026603183746)
        (135027,0.329446401500701)
        (140028,0.329283741922378)
        (145029,0.329019341535568)
        (150030,0.329194443836212)
        (155031,0.328567589950561)
        (160032,0.328264115095138)
        (165033,0.328175760736465)
        (170034,0.327864857902526)
        (175035,0.327700282497406)
        (180036,0.327831143903732)
        (185037,0.327980900936126)
        (190038,0.32828914894104)
        (195039,0.326887549400329)
        (200040,0.326607343530654)
        (205041,0.326389263029098)
        (210042,0.32631815668106)
        (215043,0.326066136856079)
        (220044,0.325944746417999)
        (225045,0.325434962577819)
        (230046,0.325219075279235)
        (235047,0.325357250585556)
        (240048,0.324935928134918)
        (245049,0.324877052869796)
        (250050,0.324749723196029)
        (255051,0.324238634281158)
        (260052,0.324024352827072)
        (265053,0.32409698097229)
        (270054,0.323453575839996)
        (275055,0.323221382894516)
        (280056,0.322972952089309)
        (285057,0.322782721691131)
        (290058,0.322951937265396)
        (295059,0.322226812086105)
        (300060,0.321884518632888)
        (305061,0.321748197431564)
        (310062,0.32149412202835)
        (315063,0.321063621702194)
        (320064,0.320820940418243)
        (325065,0.320558686132431)
        (330066,0.32068714679718)
        (335067,0.319926784715652)
        (340068,0.319712776107788)
        (345069,0.319307366056442)
        (350070,0.319022000827789)
        (355071,0.318765688867568)
        (360072,0.318396127357482)
        (365073,0.318064113006591)
        (370074,0.318199629039764)
        (375075,0.317525320730209)
        (380076,0.317121573534011)
        (385077,0.316858160848617)
        (390078,0.316555987339019)
        (395079,0.316217902297973)
        (400080,0.315862906360626)
        (405081,0.315531402463913)
        (410082,0.315236318874359)
        (415083,0.314850826883316)
        (420084,0.314562448635101)
        (425085,0.314268686361312)
        (430086,0.31395628786087)
        (435087,0.313684972400665)
        (440088,0.313422096538543)
        (445089,0.313177836456298)
        (450090,0.312949107141494)
        (455091,0.312718677892684)
        (460092,0.312518591575622)
        (465093,0.312331360082626)
        (470094,0.31218882572174)
        (475095,0.312072084169387)
        (480096,0.311986823253631)
        (485097,0.311917673139572)
        (490098,0.31188175953865)
        (495099,0.31186040356636)
        
    };
    \addlegendentry{\DFN}

    \end{axis}
\end{tikzpicture}
    \vspace{-5mm}
    \caption{\textbf{Dataset impact on pre-training performance.} On the one
    hand, pre-training using IN-1k leads to overfitting, even for the \Ours-0.6B
    model. On the other hand, pre-training using the uncurated DFN-2B dataset
    prevents overfitting but converges to a similar point due to the
    distributional shift. Pre-training on \DFN, a data mixture that predominantly
    consists of DFN-2B with a small presence of IN-1k samples leads to the best
    performance.}
    \label{fig:data_mix_loss}
\end{figure}

\begin{table}[htb!]
    \centering
    \setlength{\tabcolsep}{10pt}
    \renewcommand{\arraystretch}{1.2}
    \resizebox{0.9\linewidth}{!}{
    \begin{tabular}{lccc}
         pre-training dataset & IN-1k & DFN-2B & \DFN \\
         \shline
         \textit{attentive} & 73.5 & 74.5 & \textbf{75.6} \\
    \end{tabular}}
    \caption{\textbf{Dataset impact of downstream performance~(15 benchmarks).} The behavior in Figure~\ref{fig:data_mix_loss} is consistent with the downstream performance where we observe that using a data mixture of DFN-2B and IN-1k results in the best performance.}
    \label{tab:datset_mix_downstream}
    \vspace{-3mm}
\end{table}

\par \noindent \textbf{Compute-optimal pre-training.} Since we do not observe
signs of overfitting when we train using the \DFN dataset, we proceed to examine
the impact of extending the length of our pre-training schedule. In
\cref{fig:flops_data_params}, we study the impact of increasing the length of
the pre-training schedule from 500k to 1.2M iterations, \ie, 2B to 5B images
seen during pre-training. We observe that models pre-trained with a longer
schedule achieve significantly lower validation loss. This suggests that one can
improve the performance of \Ours either by increasing the model capacity or by
pre-training for longer schedules. Interestingly, we find that lower-capacity
models trained for a longer schedule achieve comparable validation loss to
higher-capacity models trained for a shorter schedule while using a similar
amount of FLOPs. This finding is consistent with ~\citet{hoffmann2022training}
and implies that \Ours could follow similar scaling laws. However, we defer
further investigations in this aspect for future work.

\begin{table*}[htb!]
	\centering
  	\subfloat[ \textbf{Targets}.
	\label{tab:output_type_ablation}
	]{
		\centering
		\begin{minipage}{0.4\linewidth}{\begin{center}
                \tablestyle{3pt}{1.2}
                \begin{tabular}{l  cccc}
                    target & pixels & norm. pixel~\cite{he2021masked} & KMeans &
                    dVAE~\cite{ramesh2021zero} \\
                    \shline
                    \textit{linear}    & 67.5 & \colorcell \textbf{70.0} & 66.6
                    & 64.0 \\
                    \textit{attentive} & 76.2 & \colorcell \textbf{78.2}  &
                    75.9 & 74.5 \\

                \end{tabular}
		\end{center}}\end{minipage} } \subfloat[ \textbf{Autoregression
	Pattern (causal)}.
	\label{tab:autoregressive_pattern}
	]{
		\centering
		\begin{minipage}{0.35\linewidth}{\begin{center}
                \tablestyle{3pt}{1.2}
               \begin{tabular}{l  cccc}
                    pattern & raster & spiral & checkerboard & random \\
                    \shline
                    \textit{linear}    & \colorcell \textbf{69.5} & 67.7 &  68.2
                    & 65.8 \\
                    \textit{attentive} & \colorcell \textbf{77.4} & 76.3 &  76.0
                    & 75.7 \\
                \end{tabular}
		\end{center}}\end{minipage} } \subfloat[ \textbf{Crop Scale}.
	\label{tab:crop_scale_ablation}
	]{
		\centering
		\begin{minipage}[t]{0.2\linewidth}{\begin{center}
                \tablestyle{3pt}{1.2}
                \begin{tabular}{l  ccc}
                     crop scale & 0.08  & 0.4 & 1.0 \\
                    \shline
                    \textit{linear}    & 68.4 & \colorcell \textbf{70.0} & 49.6
                    \\
                    \textit{attentive} & 77.7  & \colorcell \textbf{78.2} & 63.5
                    \\
                \end{tabular}
		\end{center}}\end{minipage} }

        \subfloat[ \textbf{Attention Structure}.
	\label{tab:prefix_ablation}
	]{
		\centering
		\begin{minipage}[t]{0.4\linewidth}{\begin{center}
                \tablestyle{3pt}{1.0}
            \begin{tabular}{l cc  cc}
                 pre-training attn. & \multicolumn{2}{c}{causal} &
                 \multicolumn{2}{c}{prefix} \\  \cmidrule{2-3} \cmidrule{4-5}
                 inference attn. & causal & bidirectional & causal &
                 bidirectional  \\
                \shline
                \textit{linear}    & 69.5 & 30.9 & 68.4 & \colorcell
                \textbf{70.0} \\
                \textit{attentive} & 77.4 & 52.3 & 76.9 & \colorcell
                \textbf{78.2} \\
             \end{tabular}
		\end{center}}\end{minipage} } \subfloat[ \textbf{Head Design}.
	\label{tab:head_design_ablation}
	]{
		\centering
		\begin{minipage}{0.35\linewidth}{\begin{center}
                \tablestyle{3pt}{1.1}
                \begin{tabular}{l  ccc}
                    head & None & MLP & Transformer  \\
                    \shline
                    \textit{linear}    & 64.0 & \colorcell 70.0 & \textbf{70.5}
                    \\
                    \textit{attentive} & 75.4 & \colorcell 78.2 & \textbf{78.5}
                    \\
                \end{tabular}
		\end{center}}\end{minipage} } \subfloat[ \textbf{Architecture}.
	\label{tab:depth_vs_width}
	]{
		\centering
		\begin{minipage}[t]{0.2\linewidth}{\begin{center}
                \tablestyle{3pt}{1.1}
                \begin{tabular}{l  cccc}
                     architecture & deep & wide  \\
                    \shline
                    \textit{linear}    & 68.8 & \colorcell \textbf{70.0} \\
                    \textit{attentive} & 77.9 & \colorcell \textbf{78.2}  \\
                \end{tabular}
		\end{center}}\end{minipage} }

    \caption{\textbf{Ablations} We investigate various design choices of \Ours. We use an \Ours-0.6B model that is pre-trained and evaluated using IN-1k. We report the linear and attentive probing results.
    The default settings for \Ours used for the main results are highlighted in \colorbox{Gray}{gray}.}    
    \label{tab:ablations}
    
 \end{table*}

\subsection{Architecture and Design}
In this section, we investigate the impact of some variations in our model and
training objective. These ablations are conducted using an \Ours-0.6B model,
which has been pre-trained and evaluated on the IN-1k dataset. The results of
these ablations are presented in Table~\ref{tab:ablations}.

\begin{figure}[t!]
    \centering
    \definecolor{CustomBlue}{HTML}{2b6774}
\definecolor{darkspringgreen}{HTML}{8ec7b7}

\begin{tikzpicture}
    \begin{axis}[
        xtick={1, 2, 4, 8},
        legend pos=south west,
        xticklabels={1$e^{21}$, 2$e^{21}$, 4$e^{21}$, 8$e^{21}$}, %
        ytick={0.292, 0.296, 0.3, 0.304, 0.308, 0.31},
        yticklabels={0.292, 0.296, 0.3, 0.304, 0.308, 0.31},
        ymax=0.3088,
        ymin=0.2915,
        xmax=11,
        xmin=0.5,
        grid=both,
        grid style={line width=.1pt, draw=gray!10},
        major grid style={line width=.2pt,draw=gray!50},
        minor tick num=2,
        axis x line*=bottom,
        axis y line*=left,
        height=2in,
        width=\linewidth,
        ylabel style= {align=center, font=\footnotesize},
        xlabel style = {font=\footnotesize},
        ylabel={Pre-training Validation Loss},
        xlabel={FLOPs \scriptsize{(log scale)}},
        yticklabel style = {font=\footnotesize},
        xticklabel style = {font=\footnotesize},
        legend style={cells={align=left}, font=\footnotesize, fill=none},
        legend cell align={center},
        xmode=log,
    ]

   \addlegendimage{mark=square*, darkspringgreen, mark options={solid}, line width=1.5pt, mark size=2.5pt, only marks}
   \addlegendimage{mark=square*, CustomBlue, mark options={solid}, line width=1.5pt, mark size=2.5pt, only marks}
    \addlegendentry{2B images}
    \addlegendentry{5B images}

   \addlegendimage{empty legend}
   \addlegendentry{}

   \addlegendimage{mark=oplus*, gray, mark options={solid}, line width=1.5pt, mark size=2pt, only marks}
   \addlegendimage{mark=oplus*, gray, mark options={solid}, line width=1.5pt, mark size=3.5pt, only marks}
   \addlegendimage{mark=oplus*, gray, mark options={solid}, line width=1.5pt, mark size=5pt, only marks}
    \addlegendentry{1B}
    \addlegendentry{3B}
    \addlegendentry{7B}

    \addplot[mark=oplus*, darkspringgreen, mark options={solid}, line width=1.5pt, mark size=3.0pt, only marks] plot coordinates {
        (0.8,  0.3084)
    };

    \addplot[mark=oplus*, darkspringgreen, mark options={solid}, line width=1.5pt, mark size=8.0pt, only marks] plot coordinates {
        (1.6, 0.3039)
    };

    \addplot[mark=oplus*, darkspringgreen, mark options={solid}, line width=1.5pt, mark size=15.0pt, only marks] plot coordinates {
        (3.51, 0.2989)
    };

    \addplot[mark=oplus*, CustomBlue, mark options={solid}, line width=1.5pt, mark size=3.0pt, only marks] plot coordinates {
        (2.07, 0.3041)
    };
    
    \addplot[mark=oplus*, CustomBlue, mark options={solid}, line width=1.5pt, mark size=8.0pt, only marks] plot coordinates {
        (4.0, 0.2991)
    };

    \addplot[mark=oplus*, CustomBlue, mark options={solid}, line width=1.5pt, mark size=15.0pt, only marks] plot coordinates {
        (8.7, 0.2945)
    };

    \end{axis}
\end{tikzpicture}
    \caption{\textbf{Scaling in FLOPs.} 
    That total number of FLOPs during training correlates with the final validation loss, suggesting compute driven scaling law similar to~\citet{hoffmann2022training}.}
    \label{fig:flops_data_params}
\end{figure}
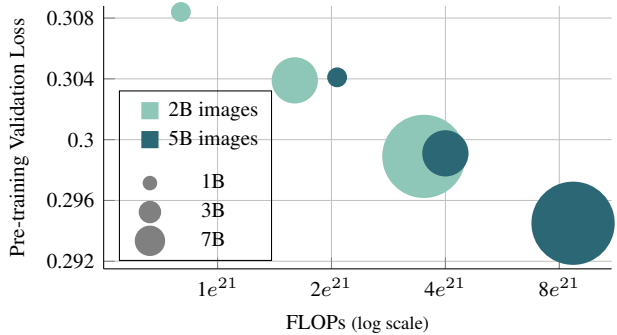 

\par \noindent \textbf{Targets and objective (a).} We explore various potential
representations for the target patches.  One approach is to utilize the raw
pixel values, and training the model with mean squared error (MSE) regression
loss. A second option, proposed by~\citet{he2021masked}, involves using
per-patch normalized pixel values instead of the raw signal with the same MSE
loss. Finally, another option is to use a discretized representation of the
patches, either using k-means or a discrete VAE~\cite{ramesh2021zero,
van2017neural}. In this case, the model is trained using a cross-entropy
objective similar to language modeling. Our experiments show that \Ours performs
best when using the MSE objective with normalized pixel values.

\par \noindent \textbf{Autoregression pattern (b).} Autoregressive pre-training
typically follows a specific order of traversal to facilitate the prediction of
the next token. In the case of language, the traversal pattern is clear, as text
is read and written one word at a time in a sequential manner (\eg, left to right
for English). However, for images, determining the traversal pattern is less
obvious. We explore various deterministic patterns, including raster, spiraling
out, checkerboard, and randomly pre-sampled patterns. Detailed examples of each
pattern are found in Appendix~\ref{app:patterns}. Even though our model performs
reasonably well with each pattern, we observe that the raster pattern leads to
significantly higher performance.

To gain deeper insights into this result, we examine the difficulty of
predicting patches along sequences for each pattern. This can be done by
measuring the loss value per patch as we progress along a sequence, as
illustrated in Figure~\ref{fig:patterns_chunks}. Our observation is that
patterns that present a more uniform distribution of difficulty across patches
result in superior models, as compared to patterns where the prediction becomes
progressively easier as the sequence unfolds. We attribute this to the
difficulty of predicting patches throughout the sequence that forces the model
to retain more information about the image. This leads to better patch features,
and consequently, to better image representation as a whole.

\par \noindent \textbf{Cropping scale (c).} We explore the impact of the
information content of each patch by adjusting the lower bound of the cropping
scale. On the one hand, opting for a cropping scale that is too small leads to
an easier next-patch-prediction task as neighboring patches' similarity
increases. On the other hand, using a large cropping scale can lead to severe
overfitting unless the dataset size is sufficiently large. Since this study is
conducted using IN-1k, we observe a clear drop in performance due to
overfitting.

\par \noindent \textbf{Causal \vs Prefix Attention (d).} We measure the impact
of incorporating prefix attention during pre-training, as opposed to using
standard causal attention. We observe that pre-training with causal
self-attention produces models that are effective in downstream transfer tasks
only when the causal mask is preserved. These models experience a significant
decline in performance when bidirectional attention is employed. However,
pre-training with prefix attention leads to models that operate effectively in
both causal and bidirectional modes. Notably, the best performance is achieved
when combining prefix attention during pre-training with bidirectional attention
during downstream adaptation.

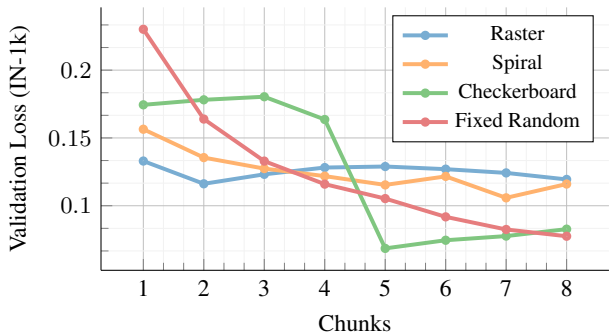
\begin{figure}[htb]
    \centering
    \definecolor{CustomOrange}{rgb}{0.8823529411,0.63725490196,0.0156862745}
\definecolor{darkspringgreen}{rgb}{0.09, 0.45, 0.27}
\definecolor{colorA}{HTML}{1F77B4} %
\definecolor{colorB}{HTML}{FF7F0E} %
\definecolor{colorC}{HTML}{2CA02C} %
\definecolor{colorD}{HTML}{D62728} %

\begin{tikzpicture}
    \begin{axis}[
        xtick={1, 2, 3, 4, 5, 6, 7, 8}, %
        legend pos=north east,
        xticklabels={1, 2, 3, 4, 5, 6, 7, 8}, %
        grid=both,
        grid style={line width=.1pt, draw=gray!10},
        major grid style={line width=.2pt,draw=gray!50},
        minor tick num=2,
        axis x line*=bottom,
        axis y line*=left,
        height=2in,
        width=\linewidth,
        ylabel style= {align=center, font=\small},
        xlabel style = {font=\small},
        ylabel={Validation Loss (IN-1k)},
        xlabel={Chunks},
        yticklabel style = {font=\small},
        xticklabel style = {font=\small},
        legend style={cells={align=left}, font=\footnotesize},
        scaled ticks=false,
    ]
    \addplot[colorA!60, mark=o, mark options={solid}, line width=1.5pt, mark size=1.0pt] plot coordinates {
        (1, 0.13296732938768388)
        (2, 0.11618377417768116)
        (3, 0.12316138552591911)
        (4, 0.12816352760103455)
        (5, 0.12892515430874443)
        (6, 0.12697348587023785)
        (7, 0.12421655586055884)
        (8, 0.11940878726814022)
    };
    \addlegendentry{Raster}
    \addplot[colorB!60, mark=o, mark options={solid}, line width=1.5pt, mark size=1.0pt] plot coordinates {
        (1, 0.15642073294823025)
        (2, 0.13550619139310258)
        (3, 0.1273931965121558)
        (4, 0.12184755553468277)
        (5, 0.11535872508962258)
        (6, 0.12163621847555536)
        (7, 0.10582898918267349)
        (8, 0.11600839086397723) 
    };
    \addlegendentry{Spiral}
    \addplot[colorC!60, mark=o, mark options={solid}, line width=1.5pt, mark size=1.0pt] plot coordinates {
        (1, 0.1744173953094472)
        (2, 0.17811322997377627)
        (3, 0.18038273246614317)
        (4, 0.16360104300540074)
        (5, 0.06855309001493191)
        (6, 0.07455928564530387)
        (7, 0.07763109998439947)
        (8, 0.08274212360059727)
    };
    \addlegendentry{Checkerboard}
    \addplot[colorD!60, mark=o, mark options={solid}, line width=1.5pt, mark size=1.0pt] plot coordinates {
        (1, 0.23013076816244782)
        (2, 0.16392436059558357)
        (3, 0.13291352050699162)
        (4, 0.11592974043725833)
        (5, 0.10528335538386194)
        (6, 0.09180085814322368)
        (7, 0.08249212394102641)
        (8, 0.07752527282960668)
    };
    \addlegendentry{Fixed Random}
    \end{axis}
\end{tikzpicture}
    \caption{\textbf{Autoregression patterns} We explore a number of patterns for
    the autoregressive traversal of an image. The set of image patches is broken
    into equal-sized chunks and the validation loss is measured per chunk. We
    observe that the way the task difficulty is distributed across chunks varies
    strongly among patterns.}
    \label{fig:patterns_chunks}
\end{figure}
\begin{table}[htb!]
	\centering
  	\subfloat[
	\textbf{MLP width}.
	]{
		\centering
		\begin{minipage}{0.5\linewidth}{\begin{center}
                \tablestyle{5pt}{1.2}
                \begin{tabular}{l ccc}
                    width & 512 & 1024 & 2048 \\
                    \shline
                    \textit{linear}    & 69.4 & 69.6 & \colorcell \textbf{70.0} \\
                    \textit{attentive} & 77.7 & 78.1 & \colorcell \textbf{78.2} \\
                \end{tabular}
		\end{center}}\end{minipage}
	}
  	\subfloat[
	\textbf{MLP depth}.
	]{
		\centering
		\begin{minipage}{0.5\linewidth}{\begin{center}
                \tablestyle{5pt}{1.2}
                \begin{tabular}{l ccc}
                    depth & 6 & 8 & 12 \\
                    \shline
                    \textit{linear}    & 65.3 & 68.1 & \colorcell \textbf{70.0} \\
                    \textit{attentive} & 76.2 & 77.1 & \colorcell \textbf{78.2} \\
                \end{tabular}
		\end{center}}\end{minipage} } \caption{\textbf{MLP design.} We vary
	the capacity of the MLP head by changing the number of MLP blocks (\ie depth)
	or the embedding size (\ie width).  Downstream
	performance improves with more capacity in either width or depth,
	but depth has more impact.}
	\label{tab:mlp_capacity}
        \vspace{-3mm}
\end{table}

\begin{table}[htb!]
    \centering
    \setlength{\tabcolsep}{9pt}
    \renewcommand{\arraystretch}{1.1}
    \resizebox{0.92\linewidth}{!}{
    \begin{tabular}{lccc}
          & \multirow{2}{*}{autoregressive} & \multicolumn{2}{c}{masked image
          modeling} \\ \cmidrule{3-4} & &  ratio=50\%  & ratio=75\% \\
         \shline
         \textit{attentive} & \textbf{78.2} & 70.3 & 77.8 \\
    \end{tabular}} \caption{\textbf{Autoregressive \vs Masking} We evaluate the
    IN-1k performance of the autoregressive objective of \Ours, in comparison to
    the masking objective~\cite{devlin2018bert,bao2021beit}. We keep all the
    other architectural and optimization components fixed. We observe that,
    under the same pre-training settings, the frozen-trunk performance of the
    autoregressive objective outperforms masking.}
    \label{tab:mim_vs_ar}
\end{table}

\begin{table*}[t!]
    \centering
    \setlength{\tabcolsep}{6pt}
    \renewcommand{\arraystretch}{1.2}
    \resizebox{1.0\linewidth}{!}{
    \begin{tabular}{llccccccccccccccccc}
        \textbf{Model} & Arch. & Data & \small{\rotatebox{90}{IN-1k}} & \small{\rotatebox{90}{iNAT-18}} & \small{\rotatebox{90}{Cifar10}}  & \small{\rotatebox{90}{Cifar100}} & \small{\rotatebox{90}{Food101}} & \small{\rotatebox{90}{DTD}} & \small{\rotatebox{90}{Pets}} & \small{\rotatebox{90}{Cars}} & \small{\rotatebox{90}{iWildCam}} & \small{\rotatebox{90}{Camelyon17}} & \small{\rotatebox{90}{PCAM}} & \small{\rotatebox{90}{RxRX1}}  & \small{\rotatebox{90}{EuroSAT}} & \small{\rotatebox{90}{fMoW}} & \small{\rotatebox{90}{Infographic}} & Avg\\
         \shline

         DINO~\cite{caron2021emerging} & ViT-B/8 & IN-1k & 80.1 & 66.0 & 97.8 & 87.3 & 89.5 & 78.4 & 92.3 & 89.2  & 58.5 & 93.7 & 90.2 & 6.1 & 98.2 & 57.0 & 41.1 & 75.0 \\
         iBOT~\cite{zhou2021ibot} & ViT-L/16 & IN-21k & 83.5 & 70.5 & 99.2 & 93.3 & 93.5 & 81.6 & 92.8 & 90.8   & 61.8 & 94.5 & 90.0 & 5.9 & 98.0 & 60.3 & 47.7 & 77.6 \\
         DINOv2~\cite{oquab2023dinov2} & ViT-g/14$_{516}$ & LVD & 86.4 & 84.5 & 99.6 & 95.2 & 96.3 & 86.3 & 96.4 & 95.6 & 68.2  & 96.5 & 90.7 & 8.0 & 98.6 & 66.7 &  58.8 & 81.9 \\
         \midrule
         BEiT~\cite{bao2021beit} & ViT-L/14 & IN-21k &
         62.2 & 44.4 & 94.4 & 78.7 & 79.0 & 64.0 & 80.9 & 69.5  & 52.0 &	92.8 & 88.2 & 4.2 &	97.5 &	47.7 & 25.9 & 65.4  \\
        \multirow{2}{*}{MAE~\cite{he2021masked, singh2023effectiveness}} & ViT-H/14 & IN-1k
        & 80.9 %
        & 64.6 %
        & 97.1
        & 85.8
        & 90.2 %
        & 78.1 %
        & 95.0 %
        & 93.7 %
        & 58.1    %
        & 94.2    %
        & 89.8    %
        & 5.4   %
        & 98.1 %
        & 56.9 %
        & 42.2
        & 75.3
        \\

        & ViT-2B/14 & IG-3B
        & 82.2 %
        & 70.8 %
        & 97.5
        & 87.3
        & 93.4 %
        & 81.2 %
        & 95.1 %
        & 94.9 %
        & 57.8    %
        & 94.4    %
        & 90.3    %
        & 7.3    %
        & 98.2 %
        & 60.1 %
        & 50.2
        & 77.4
        \\

         \midrule
         \Ours-0.6B  & ViT-H/14 & \multirow{4}{*}{\DFN} 
          & 78.5 %
          & 64.0 %
          & 97.2
          & 86.8
          & 90.1 %
          & 80.1 %
          & 93.0 %
          & 93.0 %
          & 57.9 %
          & 94.3  %
          & 90.0 %
          & 7.8 %
          & 98.4 %
          & 58.3 %
          & 45.2
          & 75.6
          \\
          \Ours-1B  & ViT-1B/14 & &
            80.6  %
          & 67.2   %
          & 98.2
          & 88.3
          & 91.6 %
          & 81.8 %
          & 93.4 %
          & 93.9  %
          & 58.6  %
          & 94.5 %
          & 90.0  %
          & 9.0 %
          & 98.6 %
          & 59.8 %
          & 47.5
          & 76.9
          \\

          \Ours-3B & ViT-3B/14 & &
            82.2 %
          & 69.7  %
          & 98.4
          & 89.9
          & 92.7 %
          & 81.9 %
          & 94.1 %
          & 93.8  %
          & 58.8 %
          & 94.3 %
          & 90.4  %
          & 9.7 %
          & 98.5 %
          & 60.9 %
          & 48.9
          & 77.6 \\

         \Ours-7B   & ViT-7B/14 & &
            82.4 %
          & 70.9  %
          & 98.6 
          & 90.0
          & 93.1 %
          & 82.3 %
          & 93.8 %
          & 92.1 %
          & 59.5 %
          & 93.6 %
          & 90.7  %
          & 10.1 %
          & 98.6 %
          & 61.7 %
          & 49.6
          & 77.8 \\

        \midrule
        
         \Ours-7B$\dagger$   & ViT-7B/14 & \DFN &
           84.0 %
          & 75.5 %
          & 98.9 
          & 91.8
          & 94.1 %
          & 85.6 %
          & 95.4 %
          & 95.0 %
          & 61.4 %
          & 94.2  %
          & 90.5 %
          & 8.4 %
          & 98.5 %
          & 63.5 %
          & 57.7
          & 79.6 \\		

    \end{tabular}}
    \caption{\textbf{Downstream evaluation with a frozen trunk.} We assess the quality of \Ours features by evaluating against a diverse set of 15 image recognition benchmarks. \Ours and the baseline methods are evaluated using attentive probing with a frozen trunk. \Ours models exhibit a strong performance across all benchmarks, especially the \Ours-7B. \Ours outperforms all other methods, using joint-embedding or generative approaches, except for DINOv2 which utilizes higher-resolution images, that typically results in a 1-1.5\% improvement on ImageNet for instance. $\dagger$: Extracting features from the 20$^\text{th}$ layer instead of the last (32$^\text{nd}$), see~\cref{tab:last_vs_best} for more details.}
    \label{tab:downstream_perf}
\end{table*}

\par \noindent \textbf{Head design (e).} We consider different types of heads on
top of the backbone to make predictions at the pixel level. Using no heads (\ie
\textit{None}) performs reasonably well, but adding an MLP further improves the
quality of the backbone. Interestingly, replacing the MLP with a full-fledged
transformer of the same depth and width only yields a marginal performance
improvement but at a significantly higher computational cost. Therefore, we opt
to use an MLP head in our approach. We hypothesize that these heads specialize
in capturing the low-level signals necessary for accurate pixel-level
prediction. By incorporating a head, the trunk can learn higher-level features
that are more suitable for downstream transfer. A similar design was employed
for contrastive learning to prevent the backbone from specializing in predicting
specific image transformations~\cite{chen2020simple}.

\par \noindent \textbf{Deeper \vs Wider architecture (f).} We present the design
specifications of \Ours in Table~\ref{tab:model_specs}, outlining its width and
depth. Unlike the original design of ViT~\cite{dosovitskiy2020image}, where the
depth is scaled more rapidly than the width, we adopt a scaling strategy similar
to that of Llama~\cite{touvron2023llama}. This allows us to scale our model more
gracefully while maintaining a reasonable depth. We validate the effectiveness
of a wider architecture in Table~\ref{tab:depth_vs_width}. Our findings indicate
that even for the relatively small-scale \Ours-0.6B model, a wider architecture
not only delivers strong performance but also improves training stability. This
observation supports the notion that some of the insights gained from training
LLMs can be similarly applied to other domains.

\par \noindent \textbf{Attentive \vs Linear probe.} For all ablations we report
the linear and attentive probing results. We observe that, consistently across
all experiments, attentive pooling provides a significant boost to performance
as it allows for a more nuanced aggregation of local features circumventing one
of the main weaknesses of generative pre-training: the absence of an image-level
global descriptor.

\par \noindent \textbf{Structure of the MLP.} The MLP plays an important role as
ablated in Table~\ref{tab:head_design_ablation}. In
Table~\ref{tab:mlp_capacity}, we further investigate the capacity of the MLP
head and how it impacts downstream performance. We vary the capacity of the head
by either changing the number of MLP blocks or their width. By default, we use a
head of 12 blocks and an embedding dimension of 2048. First, we observe that
increasing the capacity of the MLP either through depth or width leads to
consistent improvement in the downstream performance. Second, we find that
increasing the number of MLP blocks, with a fixed width, leads to a larger
improvement compared to increasing the width for a fixed depth. Interestingly,
we could not find a point where increasing the MLP capacity failed to yield
further improvements. We did not explore higher capacities beyond those reported
in Table~\ref{tab:mlp_capacity} as it would lead to models with disproportionate
head and trunk capacity.

\subsection{Pre-training objective}

\par \noindent \textbf{Autoregressive \vs Masking} We conduct a comparison
between our architecture trained with an autoregressive objective and the
masking objective popularized by BERT~\cite{devlin2018bert} for language, and by
BEiT and MAE for vision. It is important to note that we applied the masking
objective in the same setting as \Ours, thereby isolating the impact on the
performance of the pre-training objective from other design choices that differ
between \Ours and other approaches. In the masking baseline, we randomly sample
masks and replace the masked patches with learnable mask tokens.

In Table~\ref{tab:mim_vs_ar}, we show that \Ours performs better with an
autoregressive objective than a masking objective. This is consistent with the
results reported by~\citet{chen2020generative}, providing further evidence that
our improvements stem from the utilization of an autoregressive objective.

\subsection{Comparison with other methods}

In \cref{tab:downstream_perf}, we compare the attentive probing performance of
\Ours to other state-of-the-art methods across a set of 15 diverse benchmarks
that are detailed in~\cref{app:eval_benchmarks}. 

\par \noindent \textbf{Generative methods.} \Ours provides a strong performance
compared to its generative counterparts. \Ours outperforms
BEiT~\cite{bao2021beit} by a large margin. Additionally, \Ours-0.6B provides a
better performance, averaged across all benchmarks, compared to
MAE-H~\cite{he2021masked} which has an equivalent capacity. Moreover, we compare
against the MAE-2B~\cite{singh2023effectiveness} model which has been
pre-trained on IG-3B, a private dataset of 3 billion images from Instagram. We
find that both \Ours-3B and \Ours-7B outperform MAE-2B, with \Ours-7B exhibiting
a particularly large improvement. It is worth noting that, similar to \Ours, two
other generative approaches, BEiT and MAE, benefit from attentive probing,
thereby narrowing the gap between generative and joint embedding methods.

\par \noindent \textbf{Joint embedding methods.} \Ours provides a competitive
performance with joint embedding methods such as DINO~\cite{caron2021emerging},
iBOT~\cite{zhou2021ibot}, and DINOv2~\cite{oquab2023dinov2}. In terms of average
accuracy across all benchmarks, \Ours outperforms DINO and iBOT. However, it
falls behind DINOv2 which achieves its results by evaluating with
higher-resolution inputs. Note that \Ours attains such competitive performance
using higher capacity trunks. Nevertheless, \Ours's pre-training is
significantly simpler and can be trivially scaled in terms of parameters and
data, yielding consistent improvements. On the contrary, state-of-the-art joint
embedding methods like DINOv2 heavily rely on a number of tricks, such as
multi-crop augmentation, KoLeo regularization, LayerScale, Stochastic Depth,
schedules for teacher momentum and weight decay, and high-resolution fine-tuning
in order to achieve strong performance. 

\par \noindent \textbf{Extracting stronger features.} We observe that
higher-quality features can be extracted from shallower layers compared to the
last layer's features. This is likely due to the generative nature of the
pre-training objective that is inherently different than the discriminative
downstream tasks and therefore, the features with the highest semantic content
do not necessarily concentrate around the last layer.
In~\cref{tab:last_vs_best}, we report the IN-1k top-1 accuracy for features
extracted from the last layer compared to the layer with the highest
performance. A more detailed analysis of this phenomenon is provided in
Appendix~\ref{app:best_vs_last}.

\begin{table}[htb!]
    \centering
    \setlength{\tabcolsep}{9pt}
    \renewcommand{\arraystretch}{1.1}
    \resizebox{1.0\linewidth}{!}{
    \begin{tabular}{lcccc}
                            & \Ours-0.6B & \Ours-1B & \Ours-3B & \Ours-7B \\
         \shline
         \textit{last layer} & 78.5 & 80.6 & 82.2 & 82.4 \\
         \textit{best layer} & \textbf{79.4} & \textbf{82.3} & \textbf{83.3} & \textbf{84.0} \\
    \end{tabular}} \caption{\textbf{Feature extraction.} The highest quality features after \Ours pre-training typically reside in shallower layers than the last. Extracting features from earlier layers leads to a non-negligible boost to the recognition performance on IN-1k.}
    \label{tab:last_vs_best}
\end{table}

\subsection{Low-Rank Adaptation}

In addition to frozen-trunk evaluation, we examine Low-Rank Adaptation
(LoRA)~\cite{hu2021lora}, a popular and efficient finetuning method. We report
the results of LoRA fintuning of \Ours in~\cref{tab:lora_results}. We observe
that LoRA is compatible with \Ours, leading to a large boost in performance
compared to frozen-trunk evaluation. For example, \Ours-7B improves by 3.9\%
(compared to the last layer's performance) while finetuning only 0.1\% percent
of the trunk parameters.

\begin{table}[htb!]
    \centering
    \setlength{\tabcolsep}{6pt}
    \renewcommand{\arraystretch}{1.1}
    \resizebox{1.0\linewidth}{!}{
    \begin{tabular}{lcccc}
         & \Ours-0.6B & \Ours-1B & \Ours-3B & \Ours-7B \\
         \shline
         \textit{attentive} & 78.5 & 80.6 & 82.2 & 82.4 \\
         \textit{LoRA (rank=8)} & 81.0 & 83.6 & 85.5 & 86.3 \\
    \end{tabular}} \caption{\textbf{Low-rank adaptation (IN-1k).} \Ours is compatible with LoRA showing large gains compared to frozen-trunk evaluations.}
    \label{tab:lora_results}
\end{table}

\section{Discussion}

In this paper, we presented a simple and scalable method for pre-training vision
models at scale without supervision. We employed a generative autoregressive
objective during pre-training and proposed several technical contributions to
better adapt it for downstream transfer. Consequently, we observed a number of
desirable properties for our Autoregressive Image Models. First, the capacity of
our models can be effortlessly scaled to 7 billion parameters using a vanilla
transformer implementation, without resorting to stability-inducing techniques
or extensive adjustments of hyperparameters for each model scale. Second,
\Ours's performance on the pre-training task has a strong correlation with
downstream performance. Third, \Ours achieves strong performance across 15
recognition benchmarks, outperforming prior state-of-the-art methods like MAE
and significantly narrowing the gap between generative and joint embedding
pre-training approaches. Finally, we did not observe any clear signs of
saturation as we scale either in terms of parameters or data, suggesting that
there is a potential for further performance improvements with larger models
trained for even longer schedules. We hope that \Ours serves as a seed for
future research in scalable vision models that \textit{effectively leverage
uncurated datasets} without any bias towards object-centric images or strong
dependence on captions.

\paragraph{Limitations.} \Ours excels in its seamless scalability and its
effective utilization of large volumes of uncurated image data. However,
alternative methods can offer different trade-offs. MAE~\cite{he2021masked}
provides high sample efficiency and can learn good representations using a small
amount of pre-training data, reducing the risk of overfitting~\cite{el2021large}
in contrast to our approach. Contrastive
methods~\cite{oquab2023dinov2,zhou2021ibot,caron2021emerging} currently result
in stronger representations for a given model size compared to generative
approaches such as MAE and \Ours, but pose significant challenges in terms of
scalability and loss tractability due to the complexity of their objective.

\section*{Acknowledgements}

The authors would like to thank Brandon McKinzie, Samira Abnar, Preetum
Nakkiran, and Jiatao Gu for valuable feedback throughout the project. We thank
Edouard Grave and Herv\'e Jegou for their inspiring discussions during the
earlier stages of the project. We thank Marco Cuturi, James Thornton, Pierre
Ablin, and Eugene Ndiaye for their support and for many fruitful discussions
throughout the project. Finally, we would like to thank the entire Machine
Learning Research team at Apple for many helpful discussions and assistance with
infra and data.
\clearpage

{
    \small
    \bibliographystyle{ieeenat_fullname}
    \bibliography{main}
}

\clearpage
\appendix
\section{Datasets}

To assess the effectiveness and general applicability of the learned
representations by \Ours, we measure its recognition accuracy on a varied
collection of 15 benchmarks in~\cref{tab:downstream_perf}. The specifics of each
benchmark can be found in~\cref{tab:dataset_descriptions}. These benchmarks
include datasets for tasks such as fine-grained recognition, medical imaging,
satellite imagery, images in natural environments, and infographic images.

\label{app:eval_benchmarks}

\begin{table}[htb!]
    \centering
    \setlength{\tabcolsep}{8pt}
    \begin{tabular}{lrrr}
        Dataset & train & test & classes \\
         \shline
         Imagenet-1k~\cite{deng2009imagenet} & 1,281,167 & 50,000 & 1000 \\
         iNAT-18~\cite{inaturalist18}     & 437,513 & 24,426 & 8142 \\
         CIFAR-10~\cite{krizhevsky2009learning}   & 50,000 & 10,000 & 10 \\
         CIFAR-100~\cite{krizhevsky2009learning}  & 50,000 & 10,000 & 100 \\
         Food101~\cite{bossard14}     & 75,750 & 25,250 & 101 \\ 
         DTD~\cite{cimpoi14describing}       & 3,760 & 1,880 & 47 \\
         Pets~\cite{parkhi12a}      & 3,680 & 3,669 & 37 \\
         Cars~\cite{KrauseStarkDengFei-Fei_3DRR2013}  & 8,144 & 8,041 & 196 \\
         iWildCam~\cite{beery2020iwildcam}    & 129,809 & 14961 & 182 \\
         Camelyon17~\cite{bandi2018detection}  & 302,436 & 34904 & 2 \\
         PCAM~\cite{veeling2018rotation}      & 262,144 & 32768 & 2 \\
         RxRx1~\cite{taylor2019rxrx1}      & 40,612 & 9854 & 1139 \\
         EuroSAT~\cite{helber2017eurosat}     & 16,200 & 5400 & 10 \\
         fMoW~\cite{christie2018functional}       & 76,863 & 19915 & 62 \\
         Infograph~\cite{peng2019moment} & 36,023 & 15,582 & 345 \\
    \end{tabular}
    \caption{\textbf{Evaluation benchmarks.} We provide the references, the number of images in the train and test sets, and the number of categories of all the 15 recognition benchmarks used in this work.}
    \label{tab:dataset_descriptions}
\end{table}

\section{Autoregression Patterns}
\label{app:patterns}

We investigate different patterns that can be used to traverse an image during
pre-training in~\cref{tab:autoregressive_pattern}. All patterns used in this
investigation are illustrated in~\cref{fig:patterns_details}.

\begin{figure}[t!]
    \begin{subfigure}[b]{0.48\linewidth}
    \centering
        \includegraphics[width=0.9\linewidth]{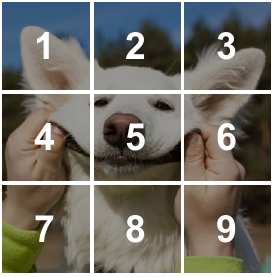}
        \caption{Raster}
    \end{subfigure}
    \begin{subfigure}[b]{0.48\linewidth}
        \centering
        \includegraphics[width=0.9\linewidth]{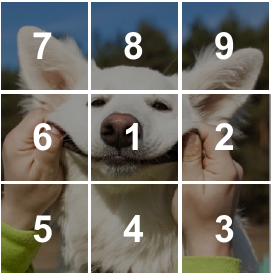}
        \caption{Spiral}
    \end{subfigure}
    \begin{subfigure}[b]{0.48\linewidth}
        \centering
        \includegraphics[width=0.9\linewidth]{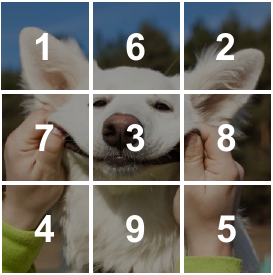}
        \caption{Checkerboard}
    \end{subfigure}
    \begin{subfigure}[b]{0.48\linewidth}
        \centering
        \includegraphics[width=0.9\linewidth]{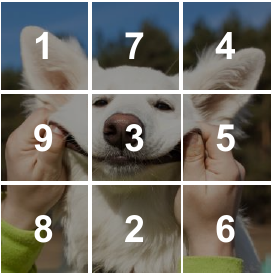}
        \caption{Random}
    \end{subfigure}
    \caption{\textbf{Autoregression patterns.} We illustrate the different autoregression patterns studied in this work including raster, spiral, checkerboard, and fixed random.
    \label{fig:patterns_details}
    }
\end{figure}

\section{Additional Analysis}
\subsection{Raster pattern validation loss}

In~\cref{fig:patterns_chunks}, we noticed that the validation loss of the raster
pattern across chunks surprisingly declined for the second chunk before
increasing again. We investigated this further in~\cref{fig:raster_rows} and
observed that this behavior is a side-effect of using the IN-1k validation set.
In particular, we observed that the top rows of the image, aside from the first
one, typically have a lower loss, whether the loss is computed over the regular
image or its vertically flipped counterpart.

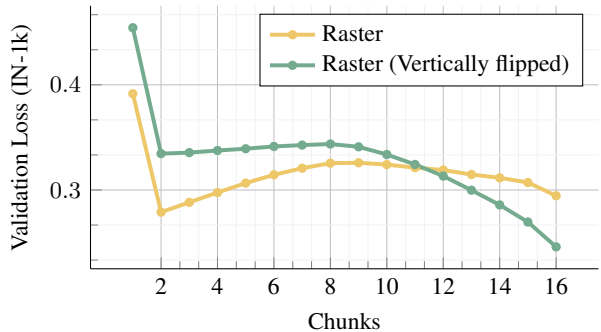
\begin{figure}
    \centering
    \definecolor{CustomOrange}{rgb}{0.8823529411,0.63725490196,0.0156862745}
\definecolor{darkspringgreen}{rgb}{0.09, 0.45, 0.27}
\definecolor{colorA}{HTML}{1F77B4} %
\definecolor{colorB}{HTML}{FF7F0E} %
\definecolor{colorC}{HTML}{2CA02C} %
\definecolor{colorD}{HTML}{D62728} %

\begin{tikzpicture}
    \begin{axis}[
        xtick={2, 4, 6, 8, 10, 12, 14, 16}, %
        legend pos=north east,
        xticklabels={2, 4, 6, 8, 10, 12, 14, 16}, %
        grid=both,
        grid style={line width=.1pt, draw=gray!10},
        major grid style={line width=.2pt,draw=gray!50},
        minor tick num=2,
        axis x line*=bottom,
        axis y line*=left,
        height=2in,
        width=\linewidth,
        ylabel style= {align=center, font=\small},
        xlabel style = {font=\small},
        ylabel={Validation Loss (IN-1k)},
        xlabel={Chunks},
        yticklabel style = {font=\small},
        xticklabel style = {font=\small},
        legend style={cells={align=left}, font=\small},
        scaled ticks=false,
        legend cell align={left},
    ]
    \addplot[CustomOrange!60, mark=o, mark options={solid}, line width=1.5pt, mark size=1.0pt] plot coordinates {
        (1, 0.3915180078125)
        (2, 0.2788911181640625)
        (3, 0.28820278076171874)
        (4, 0.29757310546875)
        (5, 0.30653154296875)
        (6, 0.3144230224609375)
        (7, 0.32068185546875)
        (8, 0.3255035571289063)
        (9, 0.32586885498046875)
        (10, 0.32415261474609375)
        (11, 0.3212524438476563)
        (12, 0.31893395751953124)
        (13, 0.314696044921875)
        (14, 0.31157844970703125)
        (15, 0.3071076708984375)
        (16, 0.2945194482421875)
    };
    \addlegendentry{Raster}
    \addplot[darkspringgreen!60, mark=o, mark options={solid}, line width=1.5pt, mark size=1.0pt] plot coordinates {
        (1, 0.4542789794921875)
        (2, 0.334581943359375)
        (3, 0.3355133837890625)
        (4, 0.3375390380859375)
        (5, 0.339215078125)
        (6, 0.3413861328125)
        (7, 0.3426744482421875)
        (8, 0.3437058544921875)
        (9, 0.34104611328125)
        (10, 0.333720185546875)
        (11, 0.3242576123046875)
        (12, 0.31323497314453125)
        (13, 0.299773916015625)
        (14, 0.2857784838867187)
        (15, 0.269506953125)
        (16, 0.24589871337890626)
    };
    \addlegendentry{Raster (Vertically flipped)}
    \end{axis}
\end{tikzpicture}
    \caption{\textbf{Raster pattern across patches.} We compute the IN-1k
    validation loss per a chunk of 16 patches (\ie, a row) for \Ours-0.6B,
    pre-trained using a raster pattern. We measure the same loss for the
    vertically flipped images of the validation set. We observe that, for IN-1k
    validation set, the patches from the top rows in the image are easier to
    predict with lower loss, likely due to the concentration of background
    patches in that region.}
    \label{fig:raster_rows}
\end{figure}

\subsection{Downstream performance across layers}

In Tables~\ref{tab:last_vs_best}~and~\ref{tab:downstream_perf}, we discussed the
gain in downstream performance that can be achieved by probing shallower layers
in the model rather than the last. We study this in more detail
in~\cref{fig:acc_across_layers}. We find that for all \Ours variants, we extract
the highest quality features, with respect to the downstream transfer, from
layers roughly at two-thirds of the way into the model depth. However, it is
important to note that the performance of deeper layers does not experience a
steep decline and continues to exhibit strong performance.
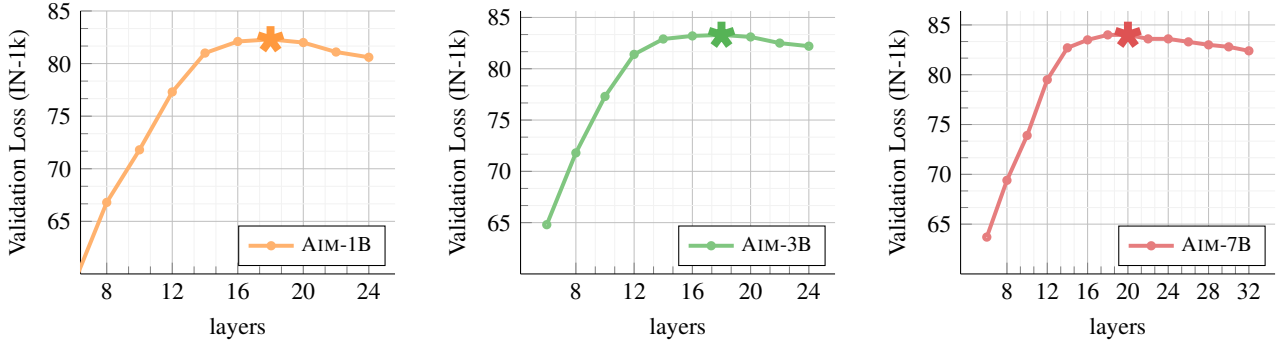
\begin{figure*}[t!]
    \begin{subfigure}[b]{0.33\linewidth}
            \definecolor{CustomOrange}{rgb}{0.8823529411,0.63725490196,0.0156862745}
\definecolor{darkspringgreen}{rgb}{0.09, 0.45, 0.27}
\definecolor{colorA}{HTML}{1F77B4} %
\definecolor{colorB}{HTML}{FF7F0E} %
\definecolor{colorC}{HTML}{2CA02C} %
\definecolor{colorD}{HTML}{D62728} %

\begin{tikzpicture}
    \begin{axis}[
        xtick={8, 12, 16, 20, 24}, %
        legend pos=south east,
        xticklabels={8, 12, 16, 20, 24}, %
        ytick={65, 70, 75, 80, 85},
        yticklabels={65, 70, 75, 80, 85},
        ymax=85.0,
        ymin=60.0,
        grid=both,
        grid style={line width=.1pt, draw=gray!10},
        major grid style={line width=.2pt,draw=gray!50},
        minor tick num=2,
        axis x line*=bottom,
        axis y line*=left,
        height=2in,
        width=\linewidth,
        ylabel style= {align=center, font=\small},
        xlabel style = {font=\small},
        ylabel={Validation Loss (IN-1k)},
        xlabel={layers},
        yticklabel style = {font=\small},
        xticklabel style = {font=\small},
        legend style={cells={align=left}, font=\footnotesize},
        scaled ticks=false,
    ]
    \addplot[colorB!60, mark=o, mark options={solid}, line width=1.5pt, mark size=1.0pt] plot coordinates {
        (6, 59.1)
        (8, 66.8)
        (10, 71.8)
        (12, 77.3)
        (14, 81.0)
        (16, 82.1)
        (18, 82.3)
        (20, 82.0)
        (22, 81.1)
        (24, 80.6)
    };
    \addlegendentry{\Ours-1B}
    \addplot[mark=star, very thick, colorB!80, mark options={mark size=5pt, solid, line width=3pt}] plot coordinates {
        (18, 82.3) 
    };
    \end{axis}
\end{tikzpicture}
    \end{subfigure}
    \hfill
    \begin{subfigure}[b]{0.33\linewidth}
            \definecolor{CustomOrange}{rgb}{0.8823529411,0.63725490196,0.0156862745}
\definecolor{darkspringgreen}{rgb}{0.09, 0.45, 0.27}
\definecolor{colorA}{HTML}{1F77B4} %
\definecolor{colorB}{HTML}{FF7F0E} %
\definecolor{colorC}{HTML}{2CA02C} %
\definecolor{colorD}{HTML}{D62728} %

\begin{tikzpicture}
    \begin{axis}[
        xtick={8, 12, 16, 20, 24}, %
        legend pos=south east,
        xticklabels={8, 12, 16, 20, 24}, %
        ytick={65, 70, 75, 80, 85},
        yticklabels={65, 70, 75, 80, 85},
        ymin=60.0,
        grid=both,
        grid style={line width=.1pt, draw=gray!10},
        major grid style={line width=.2pt,draw=gray!50},
        minor tick num=2,
        axis x line*=bottom,
        axis y line*=left,
        height=2in,
        width=\linewidth,
        ylabel style= {align=center, font=\small},
        xlabel style = {font=\small},
        ylabel={Validation Loss (IN-1k)},
        xlabel={layers},
        yticklabel style = {font=\small},
        xticklabel style = {font=\small},
        legend style={cells={align=left}, font=\footnotesize},
        scaled ticks=false,
    ]
    \addplot[colorC!60, mark=o, mark options={solid}, line width=1.5pt, mark size=1.0pt] plot coordinates {
        (6, 64.8)
        (8, 71.8)
        (10, 77.3)
        (12, 81.4)
        (14, 82.9)
        (16, 83.2)
        (18, 83.3)
        (20, 83.1)
        (22, 82.5)
        (24, 82.2)
    };
    \addlegendentry{\Ours-3B}
    \addplot[mark=star, very thick, colorC!80, mark options={mark size=5pt, solid, line width=3pt}] plot coordinates {
        (18, 83.3) 
    };
    \end{axis}
\end{tikzpicture}
    \end{subfigure}
    \begin{subfigure}[b]{0.33\linewidth}
            \definecolor{CustomOrange}{rgb}{0.8823529411,0.63725490196,0.0156862745}
\definecolor{darkspringgreen}{rgb}{0.09, 0.45, 0.27}
\definecolor{colorA}{HTML}{1F77B4} %
\definecolor{colorB}{HTML}{FF7F0E} %
\definecolor{colorC}{HTML}{2CA02C} %
\definecolor{colorD}{HTML}{D62728} %

\begin{tikzpicture}
    \begin{axis}[
        xtick={8, 12, 16, 20, 24, 28, 32}, %
        legend pos=south east,
        xticklabels={8, 12, 16, 20, 24, 28, 32}, %
        ytick={65, 70, 75, 80, 85},
        yticklabels={65, 70, 75, 80, 85},
        ymin=60.0,
        grid=both,
        grid style={line width=.1pt, draw=gray!10},
        major grid style={line width=.2pt,draw=gray!50},
        minor tick num=2,
        axis x line*=bottom,
        axis y line*=left,
        height=2in,
        width=\linewidth,
        ylabel style= {align=center, font=\small},
        xlabel style = {font=\small},
        ylabel={Validation Loss (IN-1k)},
        xlabel={layers},
        yticklabel style = {font=\small},
        xticklabel style = {font=\small},
        legend style={cells={align=left}, font=\footnotesize},
        scaled ticks=false,
    ]
    \addplot[colorD!60, mark=o, mark options={solid}, line width=1.5pt, mark size=1.0pt] plot coordinates {
        (6, 63.7)
        (8, 69.4)
        (10, 73.9)
        (12, 79.5)
        (14, 82.7)
        (16, 83.5)
        (18, 84.0)
        (20, 84.0)
        (22, 83.6)
        (24, 83.6)
        (26, 83.3)
        (28, 83.0)
        (30, 82.8)
        (32, 82.4)
    };
    \addlegendentry{\Ours-7B}
    \addplot[mark=star, very thick, colorD!80, mark options={mark size=5pt, solid, line width=3pt}] plot coordinates {
        (20, 84.0) 
    };

    \end{axis}
\end{tikzpicture}
    \end{subfigure}
    \caption{\textbf{Downstream performance across layers.} The highest quality features in terms of transfer to downstream recognition tasks can be extracted from layers different than the last, with the peak performance achieved by extracting from features roughly at two-thirds of the model depth. Deeper layers still retain a strong performance and no sharp decline is observed.
    \label{fig:acc_across_layers}
    }
\end{figure*}

\section{Hyperparameters}
\label{app:hparams}
\begin{table}[htb!]
    \begin{center}
        \centering
        \setlength{\tabcolsep}{10pt}
        \resizebox{\linewidth}{!}{
        \begin{tabular}{l|l}
            config & value \\
            \shline
            Optimizer & AdamW~\cite{loshchilov2017decoupled} \\
            Optimizer Momentum & $\beta_1=0.9,\beta_2=0.95$ \\
            Peak learning rate & $1e^{-3}$\\
            Minimum Learning rate & 0.0 \\
            Weight decay & 0.05 \\
            Batch size & 4096 \\
            Patch size & (14, 14) \\
            Gradient clipping & 1.0 \\
            Warmup iterations & 31,250 \\
            Total iterations & 1,250,000 \\
            Learning rate schedule & cosine decay~\cite{loshchilov2016sgdr} \\
            Augmentations: \\
            \quad {\tt RandomResizedCrop} \\
            \qquad {\tt size} & 224px \\
            \qquad {\tt scale} & [0.4, 1.0] \\
            \qquad {\tt ratio} & [0.75, 1.33] \\
            \qquad {\tt interpolation} & \texttt{Bicubic} \\
            \quad {\tt RandomHorizontalFlip} & $p=0.5$ \\
        \end{tabular}}
    \end{center}
    \caption{\textbf{Pre-training hyperparameters} All \Ours variants of different capacities have been trained using the same set of hyperparameters detailed above. }
    \label{tab:pretrain_settings}
    \end{table}
\begin{table}[t!]
    \begin{center}
        \centering
        \setlength{\tabcolsep}{4pt}
        \resizebox{\linewidth}{!}{
        \begin{tabular}{l|rr}
            config & IN-1k & Others \\
            \shline
            Optimizer & \multicolumn{2}{c}{AdamW~\cite{loshchilov2017decoupled}} \\
            Optimizer Momentum & \multicolumn{2}{c}{$\beta_1=0.9,\beta_2=0.999$} \\
            Peak learning rate grid & \multicolumn{2}{c}{[1, 3, 5, 10, 15, 20, 40]~$\times1e^{-4}$} \\
            Minimum Learning rate & \multicolumn{2}{c}{$1e^{-5}$} \\
            Weight decay & \multicolumn{2}{c}{0.1} \\
            Batch size & 1024 & 512 \\
            Gradient clipping & \multicolumn{2}{c}{3.0} \\
            Warmup epochs & 5 & 0 \\
            Epochs & 50 & 100  \\
            Learning rate schedule & \multicolumn{2}{c}{cosine decay~\cite{loshchilov2016sgdr}} \\
            Augmentations: \\
            \quad {\tt RandomResizedCrop} \\
            \qquad {\tt size} & \multicolumn{2}{c}{224px} \\
            \qquad {\tt scale} & \multicolumn{2}{c}{[0.08, 1.0]} \\
            \qquad {\tt ratio} & \multicolumn{2}{c}{[0.75, 1.33]} \\
            \qquad {\tt interpolation} & \multicolumn{2}{c}{\texttt{Bicubic}} \\
            \quad {\tt RandomHorizontalFlip} & \multicolumn{2}{c}{$p=0.5$} \\
            \quad {\tt Color Jitter} & \multicolumn{2}{c}{0.3} \\
            \quad {\tt AutoAugment~\cite{cubuk2019autoaugment}} & \multicolumn{2}{c}{\texttt{rand-m9-mstd0.5-inc1}} \\
        \end{tabular}}
    \end{center}
    \caption{\textbf{Attentive probe hyperparameters.} We detail the hyperparameters used for attentive probing \Ours as well as the baselines. For all experiments, we search over different learning rate values and report the best for both \Ours and baselines.}

    \label{tab:probe_settings}
    \end{table}

\par \noindent \textbf{Pre-training.} \Ours models of all capacities have been
pre-trained using the same set of hyperparameters that are reported
in~\cref{tab:pretrain_settings}. The \Ours-0.6 model however has been trained
only for the shorter schedule of 500k iterations. We did not observe any
instability while scaling the capacity of our model, thereby not requiring any
further tuning of the optimization hyperparameters.

\par \noindent \textbf{Attentive Probing.} Downstream evaluation for \Ours and
the baselines has been primarily conducted via attentive probing as described
in~\cref{sec:approach}. We report the hyperparameters used to probe all methods
in~\cref{tab:probe_settings}. For a fair comparison with other baselines, we
search over different values for the learning rate and report the best
performance of each method similar to~\cite{oquab2023dinov2}. For \Ours and
other generative baselines, we average the features for the last 6 layers of the
model before feeding them to the attention-probing head which leads to a modest
gain in performance. Note that the descriptor dimensionality remains the same
which is different from the practice of concatenating features similar to
iGPT\cite{chen2020generative} which indirectly inflates the capacity of the
evaluation head.

\par \noindent \textbf{Low-rank adaptation.} For LoRA finetuning, we use the
same hyperparameters as reported in~\cref{tab:probe_settings} in addition to
mixup~\cite{zhang2017mixup} (\texttt{alpha}=0.8). We apply LoRA adaptation, with
\texttt{rank}=8,  only to the parameters of the attention block. In particular,
the weight matrices for the queries, values, and output projection.
\label{app:best_vs_last}

\end{document}